\documentclass{article}

\usepackage{amsmath,amssymb}
\usepackage[margin = 1.5in]{geometry}
\usepackage{graphicx}
\usepackage{natbib}
\usepackage{bm}
\usepackage{float}
\usepackage{verbatim}
\usepackage{algorithm,algorithmic}
\usepackage{enumerate}
\usepackage[labelfont=bf]{caption}
\usepackage{multirow}

\newtheorem{theorem}{Theorem}
\newtheorem{lemma}{Lemma}

\newcommand{\indist}{\overset{d}{\rightarrow}}     
\newcommand{\inprob}{\overset{P}{\rightarrow}}    
\newcommand{\iid}{\stackrel{iid}{\sim}}		      
\newcommand{\ukm}{U_{n,k_n,m_n}}
\newcommand{\uk}{U_{n,k_n}}
\newcommand{\hatukm}{\hat{U}_{n,k_n,m_n}}

\newcommand{\xred}{\bm{X}^{(R)}}

\newcommand{\predpoint}{\bm{x}^*}
\newcommand{\treefn}{T}
\newcommand{\featvect}{\bm{x}_i}
\newcommand{\predpoints}{\bm{x}_{\text{\tiny TEST}}}
\newcommand{\randparam}{\omega}

\newcommand{\featspace}{\mathcal{X}}

\newcommand{\gred}{g^{(R)}}
\newcommand{\gredhat}{\hat{g}^{(R)}}
\newcommand{\nxone}{n_{\tilde{\bm{z}}}}
\newcommand{\nmc}{n_{MC}}
\newcommand{\rf}{r_{n,k_n,m_n}}

\newcommand{\randkern}{h_{k_n}^{(\omega_i)}}

\allowdisplaybreaks[4]

\begin{document}

\title{\vspace{-15mm} Quantifying Uncertainty in Random Forests via \\ Confidence Intervals and Hypothesis Tests}

\vspace{20mm}

\date{\today}
\author{{\bf Lucas Mentch} \hspace{45mm} \footnotesize{LKM54@CORNELL.EDU} \\ 
{\bf Giles Hooker} \hspace{45mm} \footnotesize{GJH27@CORNELL.EDU} \\ 
\\
\\
Department of Statistical Science \\
Cornell University \\
Ithaca, NY  14850, USA
}

\maketitle


\begin{abstract}
This work develops formal statistical inference procedures for predictions generated by supervised learning ensembles.  Ensemble methods based on bootstrapping, such as bagging and random forests, have improved the predictive accuracy of individual trees, but fail to provide a framework in which distributional results can be easily determined.  Instead of aggregating full bootstrap samples, we consider predicting by averaging over trees built on subsamples of the training set and demonstrate that the resulting estimator takes the form of a U-statistic.  As such, predictions for individual feature vectors are asymptotically normal, allowing for confidence intervals to accompany predictions.  In practice, a subset of subsamples is used for computational speed; here our estimators take the form of incomplete U-statistics and equivalent results are derived.  We further demonstrate that this setup provides a framework for testing the significance of features.  Moreover, the internal estimation method we develop allows us to estimate the variance parameters and perform these inference procedures at no additional computational cost.  Simulations and illustrations on a real dataset are provided.
\end{abstract}


\section{Introduction}
\label{sec:Introduction}

This paper develops tools for performing formal statistical inference for predictions generated by a broad class of methods developed under the algorithmic framework of data analysis. In particular, we focus on ensemble methods -- combinations of many individual, frequently tree-based, prediction functions -- which have played an important role. We present a variant of bagging and random forests, both initially introduced by \cite{bagging,randomforests}, in which base learners are built on randomly chosen subsamples of the training data and the final prediction is taken as the average over the individual outputs.  We demonstrate that this fits into the statistical framework of U-statistics, which were shown to have minimum variance by \cite{Halmos1946} and later demonstrated to be asymptotically normal by \cite{HoeffdingUstat}.  This allows us to demonstrate that under weak regularity conditions, predictions generated by these subsample ensemble methods are asymptotically normal.  We also provide a method to consistently estimate the variance in the limiting distribution without increasing the computational cost so that we may produce confidence intervals and formally test feature significance in practice.  Though not the focus of this paper, it is worth noting that this subbagging procedure -- suggested by \cite{subbagging} for use in model selection -- was shown by \cite{subbaggingrate} to outperform traditional bagging in many situations.

We consider a general supervised learning framework in which an outcome $Y \in \mathbb{R}$ is predicted as a function of $d$ features $\bm{X} = (X_1, ..., X_d)$ by the function $\mathbb{E}[Y|\bm{X}] = F(\bm{X})$.  We also allow binary classification so long as the model predicts the probability of success, as opposed to a majority vote, so that the prediction remains real valued.  Additionally, we assume a training set $\{(\bm{X}_1,Y_1), ..., (\bm{X}_n,Y_n)\}$ consisting of $n$ independent examples from the process that is used to produce the prediction function $\hat{F}$.   Throughout the remainder of this paper, we implicitly assume that the dimension of the feature space $d$ remains fixed, though nothing in the theory provided prohibits a growing number of features so long as our other explicit conditions on the statistical behavior of trees are met.


Statistical inference proceeds by asking the counterfactual question, ``What would our results look like if we regenerated these data."  That is, if a new training set was generated and we reproduced $\hat{F}$, how different might we expect the predictions to be?  To illustrate, consider the hypothesis that the feature $X_1$ does not contribute to the outcome at any point in the feature space:

\[
H_0:  \exists \hspace{1mm} F_1 \hspace{2mm} s.t. \hspace{2mm} F(x_1, ..., x_d) = F_1(x_2, ..., x_d)  \hspace{5mm} \forall \hspace{1mm} (x_1, ..., x_d) \in \featspace
\]

A formal statistical test begins by calculating a test statistic $t_0 = t((\bm{X}_1,Y_1),\ldots,(\bm{X}_n,Y_n))$ and asks, ``If our data was generated according to $H_0$ and we generated a new training set and recalculated $t$, what is the probability that this new statistic would be larger than $t_0$?''  That is, we are interested in estimating $P(t > t_0|H_0)$.  In most fields, a probability of less than 0.05 is considered sufficient evidence to reject the assumption that the data was generated according to $H_0$.  Of course, a 0.05 chance can be obtained by many methods (tossing a biassed coin, for example) so we also seek a statistic $t$ such that when $H_0$ is false, we are likely to reject.  This probability of correctly rejecting $H_0$ is known as the power of the test, with more powerful tests clearly being more useful. 

Here we propose to conduct the above test by comparing predictions generated by $\hat{F}$ and $\hat{F_1}$. Before doing so however, we consider the simpler hypothesis involving the value of a prediction:

\[
H'_0:  F(x_1,\ldots,x_p) = f_0.
\]

Though often of less scientific importance, hypotheses of this form allow us to generate confidence intervals for predictions. These intervals are defined to be those values of $f_0$ for which we do not have enough evidence to reject $H'_0$.  In practice, we choose $f_0$ to be the prediction $\hat{F}(x_1, ..., x_p)$ generated by the ensemble method in order to provide a formalized notion of plausible values of the prediction, which is, of course, of genuine interest. Our results begin here because the statistical machinery we develop will provide a distribution for the values of the prediction.  This allows us to address $H'_0$, after which we can combine these tests to address hypotheses like $H_0$.

Although this form of statistical analysis is ubiquitous in scientific literature, it is worthwhile contrasting this form of analysis with an alternative based on probably approximately correct (PAC) theory, as developed by \cite{VC} and \cite{Valiant1984}. PAC theory provides a uniform bound on the difference between the true error and observed training error of a particular estimator, also referred to as a hypothesis.  In this framework, $err(F)$ is some error of the function $F$ which is estimated by $\widehat{err}(F)$ based on the data.  A bound is then found for $P( \sup_{F \in \mathcal{F}} |\widehat{err}(F) - err(F)| > \epsilon)$ where $\mathcal{F}$ is some class of functions that includes $\hat{F}$. Since this bound is uniform over $\mathcal{F}$, it applies to $\hat{F}$ and we might think of comparing $\widehat{err}(\hat{F})$ with $\widehat{err}(\hat{F}_1)$ using such bounds. While appealing, these bounds provide the accuracy of our estimate of $err(\hat{F})$ but do not account for how the true $err(\hat{F})$ might change when $\hat{F}$ is reproduced with new training data.  The uniformity of these bounds could be used to account for the uncertainty in $\hat{F}$ if it is chosen to minimize $\widehat{err}(F)$ over $\mathcal{F}$, but this is not always the case, for example, when using tree-based methods. We also expect the same uniformity to make PAC bounds conservative, thereby resulting in tests with lower power than those we develop.

Our analysis relies on the structure of subsample-based ensemble methods, specifically making use of classic $U$-statistic theory.  These estimators have a long history (see, for example, original work by \cite{KendallsTau} or \cite{Wilcoxon}, or \cite{Lee1990} which has a modern overview), frequently focussed on rank-based non-parametric tests, and have been shown to have an asymptotically normal sampling distribution by \cite{HoeffdingUstat}.  Our application to subsample ensembles requires the extension of these results to some new cases as well as methods to estimate the asymptotic variance, both of which we provide.  


U-statistics have traditionally been employed in the context of statistical parameter estimation.  From this classical statistical perspective, we treat ensemble-tree methods like bagging and random forests as estimators and thus the limiting distributions and inference procedures we develop are with respect to the expected prediction generated by the ensemble.  That is, given a particular prediction point $\predpoint$, our limiting normal distributions are centered at the expected ensemble-based prediction at $\predpoint$ and not necessarily $F(\predpoint)$.  Such forms of distributional analysis are common in other nonparametric regression settings -- see \cite{EubankNonparametric} Section 4.8, for example.  More details on appropriate interpretations of the results are provided throughout the paper, in particular in Section 4.1.

In order to claim that the inference procedures proposed here are asymptotically valid for $F(\predpoint)$, the ensemble must consistently predict $F(\predpoint)$ at a rate of $\sqrt{n}$ or faster.  Though this is the case for many classical estimators, establishing fast uniform rates of convergence for tree-based ensembles has proven extremely difficult.  \cite{CART} discuss consistency of general partition-type models in the final chapter of their seminal book in the context of both classification and regression.  \cite{biau08} restrict their attention to classification, but prove consistency of certain idealized bagging and random forest estimators, provided the individual trees are consistent.  This paper also discusses a more general version of bagging, where the samples used to construct individual base learners may be proper subsamples of the training set taken with replacement as opposed to full bootstrap samples, so as to include the subbagging approach.  \cite{biau12} further examines the consistency of random forests and investigates their behavior in the presence of a sparse feature space.  Recently, \cite{NarrowingtheGap} proved consistency for a mathematically tractable variant of random forests and in some cases, achieved empirical performance on par with the original random forest procedure suggested by Breiman.  \cite{RLT} prove consistency for their Reinforcement Learning Trees, where embedded random forests are used to decide splitting variables, and achieve significant improvements in empirical MSE for some datasets.  However, no rates of convergence have been developed that could be applied to analyze the ensemble methods we consider here.

Beyond these consistency efforts, mathematical analyses of ensemble learners has been somewhat limited.  \cite{SextonLaake} propose estimating the standard error of bagged trees and random forests using jackknife and bootstrap estimators.  Recently, \cite{WagerIJ} proposed applying the jackknife and infinitesimal jackknife procedures introduced by \cite{Efron2013} for estimating standard errors in random forest predictions.  \cite{BART} have received significant attention for developing {\em BART}, a Bayesian ``sum-of-trees" statistical model for the underlying regression function that allows for pointwise posterior inference throughout the feature space as well as estimates for individual feature effects.  Recently, \cite{BARText} extended the BART approach by suggesting a permutation-based approach for determining feature relevance and by introducing a procedure to allow variable importance information to be reflected in the prior.

The layout of this paper is as follows:  we demonstrate in Section~\ref{sec:TreesAsUStatistics} that ensemble methods based on subsampling can be viewed as U-statistics.  In Section~\ref{sec:EstimatingTheVariance} we provide consistent estimators of the limiting variance parameters so that inference may be carried out in practice.  Inference procedures, including a test of significance for features, are discussed in Section~\ref{sec:InferenceProcedures}.  Simulations illustrating the limiting distributions and inference procedures are provided in Section~\ref{sec:Simulations} and the inference procedures are applied to a real dataset provided by Cornell University's Lab of Ornithology in Section~\ref{sec:RealData}.


\section{Ensemble Methods as U-statistics}
\label{sec:TreesAsUStatistics}

We begin by introducing the subbagging and subsampled random forest procedures that result in estimators in the form of U-statistics.  In both cases, we provide an algorithm to make the procedure explicit.


\subsection{Subbagging}
We begin with a brief introduction to U-statistics; see \cite{Lee1990} for a more thorough treatment.  Let $Z_1, ..., Z_n \iid F_{Z,\theta}$ where $\theta$ is the parameter of interest and suppose that there exists an unbiased estimator $h$ of $\theta$ that is a function of $k \leq n$ arguments.  Then we can write 

\[
\theta = \mathbb{E}h(Z_1, ..., Z_k)
\]

\noindent and without loss of generality, we may further assume that $h$ is permutation symmetric in its arguments since any given $h$ may be replaced by an equivalent permutation symmetric version.  The minimum variance unbiased estimator for $\theta$ is given by

\begin{equation}
\label{UstatDefn}
U_n = \frac{1}{\binom{n}{k}} \sum_{(i)} h(Z_{i_{1}}, ..., Z_{i_{k}})
\end{equation}

\noindent where the sum is taken over all $\binom{n}{k}$ subsamples of size $k$ and is referred to as a U-statistic with kernel $h$ of rank $k$.  When both the kernel and rank remain fixed, \cite{HoeffdingUstat} showed that these statistics are asymptotically normal with limiting variance $\frac{k^2}{n} \zeta_{1,k}$ where 

\begin{equation}
\label{ZetaDefn}
\zeta_{1,k} = cov(h(Z_1, ..., Z_k), h(Z_1, Z_{2}^{'}, ..., Z_{k}^{'}))
\end{equation}

\noindent and $Z_{2}^{'}, ..., Z_{k}^{'} \iid F_{Z,\theta}$.  The 1 in the subscript comes from the fact that there is 1 example in common between the two subsamples.  In general, $\zeta_{c,k}$ denotes a covariance in the form of (\ref{ZetaDefn}) with $c$ examples in common.

Given infinite computing power and a consistent estimate of $\zeta_{1,k}$, Hoeffding's original result is enough to produce a subbagging procedure with asymptotically normal predictions.  Suppose that as our training set, we observe $Z_1 = (\bm{X}_1,Y_1), ..., Z_n = (\bm{X}_n,Y_n) \iid F_{\bm{X},Y}$ where $\bm{X} = (X_1, ..., X_d)$ is a vector of features and $Y \in \mathbb{R}$ is the response.  Fix $k \leq n$ and let $(\bm{X}_{i_1},Y_{i_1}), ..., (\bm{X}_{i_{k}},Y_{i_k})$ be a subsample of the training set.  Given a feature vector $\predpoint \in \featspace$ where we are interested in making a prediction, we can write the prediction at $\predpoint$ generated by a tree that was built using the subsample $(\bm{X}_{i_1},Y_{i_1}), ..., (\bm{X}_{i_{k}},Y_{i_k})$ as a function $\treefn_{\predpoint}$ from $(\featspace \times \mathbb{R}) \times \cdots \times (\featspace \times \mathbb{R})$ to $\mathbb{R}$.  Taking all $\binom{n}{k}$ subsamples, building a tree and predicting at $\predpoint$ with each, we can write our final subbagged prediction at $\predpoint$ as

\begin{equation}
\label{TreeEst}
b_n(\predpoint) = \frac{1}{\binom{n}{k}} \sum_{(i)} \treefn_{\predpoint} ((\bm{X}_{i_1},Y_{i_1}), ..., (\bm{X}_{i_{k}},Y_{i_k})).
\end{equation}

\noindent by averaging the $\binom{n}{k}$ tree-based predictions.  Treating each ordered pair as one of $k$ inputs into the function $\treefn_{\predpoint}$, the estimator in (\ref{TreeEst}) is in the form of a U-statistic since tree-based estimators produce the same predictions independent of the order of the training data.  Thus, provided the distribution of predictions at $\predpoint$ has a finite second moment and $\zeta_{1,k} > 0$, the distribution of subbagged predictions at $\predpoint$ is asymptotically normal.  Note that in this context, $\zeta_{1,k}$ is the covariance between predictions at $\predpoint$ generated by trees trained on datasets with 1 sample in common.

Of course, building $\binom{n}{k}$ trees is compuationally infeasible for even moderately sized training sets and an obvious substantial improvement in computationally efficiency can be achieved by building and averaging over only $m_n < \binom{n}{k}$ trees.  In this case, the estimator in (\ref{TreeEst}), appropriately scaled, is called an \emph{incomplete} U-statistic.  When the $m_n$ subsamples are selected uniformly at random with replacement from the $\binom{n}{k}$ possibilities, the resulting incomplete U-statistic remains asymptotically normal; see \cite{Janson1984} or \cite{Lee1990} page 200 for details.

Though more computationally efficient, there remains a major shortcomming with this approach:  the number of samples used to build each tree, $k$, remains fixed as $n \rightarrow \infty$.  We would instead like $k$ to grow with $n$ so that trees can be grown to a greater depth, thereby presumably producing more accurate predictions.  Incorporating this, our estimator becomes

\begin{equation}
b_{n,k_n,m_n}(\predpoint) = \frac{1}{m_{n}} \sum_{(i)} \treefn_{\predpoint,k_n} ((\bm{X}_{i_1},Y_{i_1}), ..., (\bm{X}_{i_{k_n}},Y_{i_{k_n}})).
\label{TreeEstGrowingRank}
\end{equation}

Statistics of this form were discussed by \cite{ious} and called \emph{Infinite Order} U-statistics (IOUS) in the complete case, when $m_n = \binom{n}{k_n}$, and \emph{resampled} statistics in the incomplete case.  Specifically, Frees considers the situation where, given an $i.i.d$ sample $Z_1, Z_2, ...$ and kernel $h_{k_n}$, $\lim_{n \rightarrow \infty} h_{k_n} (Z_{i_1}, ..., Z_{i_{k_n}}) = h (Z_1, Z_2, ...)$ and $\theta = \mathbb{E} h (Z_1, Z_2, ...)$ and goes on to develop sufficient conditions for consistency and asymptotic normality whenever $m_n$ grows faster than $n$.  In contrast, the theorem below introduces a central limit theorem for estimators of the same form as in (\ref{TreeEstGrowingRank}) but with respect to their individual means $\mathbb{E}b_{n,k_n,m_n}(\predpoint)$ and covers all possible growth rates of $m_n$ with respect to $n$.  In this context, only minimal regularity conditions are required for asymptotic normality.  We begin with an assumption on the distribution of estimates for the general U-statistic case.


\vspace{5mm}
\noindent \textbf{Condition 1:  }  \emph{Let $Z_1, Z_2, ... \iid F_{Z}$ with $\theta_{k_n} = \mathbb{E}h_{k_n}(Z_1, ...,Z_{k_n})$ and define $h_{1,k_n}(z) = \mathbb{E}h_{k_n}(z,Z_2, ..., Z_{k_n}) - \theta_{k_n}$.  Then for all $\delta > 0$,}

\[
\lim_{n \rightarrow \infty} \frac{1}{\zeta_{1,k_n}} \int_{|h_{1,k_n}(Z_1)| \geq \delta \sqrt{n \zeta_{1,k_n}}} h_{1,k_n}^{2}(Z_1) dP = 0.
\]
\vspace{5mm}

\noindent This condition serves to control the tail behavior of the predictions and allows us to satisfy the Lindeberg condition needed to obtain part \emph{(i)} of Theorem 1 below.



\begin{theorem}
\label{subbaggingthm}
Let $Z_1, Z_2, ... \iid F_Z$ and let $U_{n,k_n,m_n}$ be an incomplete, infinite order U-statistic with kernel $h_{k_n}$ that satisfies Condition 1.  Let $\theta_{k_n} = \mathbb{E}h_{k_n}(Z_1, ..., Z_{k_n})$ such that $\mathbb{E}h_{k_n}^{2}(Z_1, ..., Z_{k_n}) \leq C < \infty$ for all $n$ and some constant $C$, and let $\lim \frac{n}{m_n} = \alpha$.  Then as long as $\lim \frac{k_n}{\sqrt{n}} = 0$ and $\lim \zeta_{1,k_n} \neq 0$,
\begin{enumerate}[(i)]
\item if $\alpha = 0$, then $\frac{\sqrt{n}(U_{n,k_n,m_n} - \theta_{k_n})}{\sqrt{k_{n}^{2} \zeta_{1,k_n}}} \indist \mathcal{N}(0,1)$.
\item if $0 < \alpha < \infty$, then $\frac{\sqrt{m_n}(U_{n,k_n,m_n} - \theta_{k_n})}{\sqrt{\frac{k_{n}^{2}}{\alpha} \zeta_{1,k_n} + \zeta_{k_n,k_n}}} \indist \mathcal{N}(0,1)$.
\item if $\alpha = \infty$, then $\frac{\sqrt{m_n}(U_{n,k_n,m_n} - \theta_{k_n})}{\sqrt{\zeta_{k_n,k_n}}} \indist \mathcal{N}(0,1)$.
\end{enumerate}
\label{subbaggingthm}
\end{theorem}

\noindent Condition 1, though necessary for the general U-statistic setting, is a bit obscure.  However, in our regression context, when the regression function is bounded and the errors have exponential tails, a more intuitive Lipschitz-type condition given in Proposition 1 is sufficient.  Though stronger than necessary, this alternative condition allows us to satisfy the Lindeberg condition and is reasonable to expect of any supervised learning method.  

\vspace{5mm}
\noindent \textbf{Proposition 1:  }  \emph{For a bounded regression function $F$, if there exists a constant $c$ such that for all $k_n \geq 1$, }
\begin{align*}
\big| h((\bm{X}_{1},Y_{1}), ..., (\bm{X}_{k_n},Y_{k_n}), (\bm{X}_{k_n+1},Y_{k_n+1})) - h((\bm{X}_{1},Y_{1}), ..., (\bm{X}_{k_n},&Y_{k_n}), (\bm{X}_{k_n+1},Y^{*}_{k_n+1})) \big| \\ 
\leq c \big| Y_{k_n+1} - Y^{*}_{k_n+1} \big|
\end{align*}
\noindent \emph{where $Y_{k_n+1} = F(\bm{X}_{k_n+1}) + \epsilon_{k_n+1}$, $Y^{*}_{k_n+1} = F(\bm{X}_{k_n+1}) + \epsilon^{*}_{k_n+1}$, and where $\epsilon_{k_n+1}$ and $\epsilon^{*}_{k_n+1}$ are i.i.d. with exponential tails, then Condition 1 is satisfied.}
\vspace{5mm}

A number of important aspects of these results are worth pointing out.  First, note from Theorem \ref{subbaggingthm} that the trees are built with subsamples that are approximately square root of the size of the full training set.  This condition is not necessary for the proof, but ensures that the variance of the U-statistic in part $(i)$ converges to 0 as is typically the case in central limit theorems.  By maintaining this relatively small subsample size, we can build many more trees and maintain a procedure that is computationally equivalent to traditional bagging based on full bootstrap samples.  Also note that no particular assumptions are placed on the dimension $d$ of the feature space; the number of features may grow with $n$ so long as the stated conditions remain satisfied.

The final condition of Theorem \ref{subbaggingthm}, that $\lim \zeta_{1,k_n} \neq 0$, though not explicitly controllable, should be easily satisfied in many cases.  As an example, suppose that the terminal node size is bounded by $\mathcal{T}$ so that trees built with larger training sets are grown to greater depths.  Then if the form of the response is $Y = F(\bm{X}) + \epsilon$ where $\epsilon$ has variance $\sigma^2$, $\zeta_{1,k_n}$ will be bounded below by $\sigma^2 / \mathcal{T}$.  Finally, note that the assumption of exponential tails on the distribution of regression errors in Proposition 1 is stronger than necessary.  Indeed, so long as $k_n=o(\sqrt{n})$, we need only insist that $n P \left( \big| \epsilon \big| > \sqrt{n} \right) \rightarrow 0$.


The proofs of Theorem and Proposition \ref{subbaggingthm} are provided in Appendix A.  The subbagging algorithm that produces asymptotically normal predictions at each point in the feature space is provided in Algorithm \ref{algo:subbagging}.  


\begin{algorithm}
\begin{algorithmic}

\STATE Load training set 
\STATE Select size of subsamples $k_n$ and number of subsamples $m_n$

\FOR {$i$ in 1 to $m_n$}
	\STATE Take subsample of size $k_n$ from training set
	\STATE Build tree using subsample
	\STATE Use tree to predict at $\predpoint$
\ENDFOR

\STATE Average the $m_n$ predictions to get final estimate $b_{n,k_n,m_n}(\predpoint)$
\end{algorithmic}
\caption{Subbagging}
\label{algo:subbagging}
\end{algorithm}

Note that this procedure is precisely the original bagging algorithm suggested by Breiman,  but with proper subsamples used to build trees instead of full bootstrap samples.  In Section \ref{sec:EstimatingTheVariance}, we provide consistent estimators for the limiting variance parameters in Theorem \ref{subbaggingthm} so that we may carry out inference in practice.  

We would also like to acknowledge similar work currently in progress by \cite{Wager2014}.  Wager builds upon the potential nearest neighbor framework introduced by \cite{LinJeon2006} and seeks to provide a limiting distribution for the case where many trees are used in the ensemble, roughly corresponding to our result $(i)$ in Theorems 1 and 2.  The author considers only an idealized class of trees based on the assumptions in \cite{Meinshausen2006} as well as additional \emph{honesty} and \emph{regularity} conditions that allow $k_n$ to grow at a faster rate, and demonstrates that when many Monte Carlo samples are employed, the infinitesimal jackknife estimator of variance is consistent and predictions are asymptotically normal.  This estimator has roughly the same computational complexity as those we propose in Section 3 and should scale well subject to some additional bookkeeping.  In contrast, the theory we provide here takes into account all possible rates of Monte Carlo sampling via the three cases discussed in Theorems 1 and 2 and we provide a consistent means for estimating each corresponding variance.  


\subsection{Random Forests}
\label{subsec:rf}

The distributional results described above for subbagging do not insist on a particular tree building method.  So long as the trees generate predictions that satisfy minimal regularity conditions, the experimenter is free to use whichever building method is preferred.  The subbagging procedure does, however, require that each tree in the ensemble is built according to the same method.

This insistence on a uniform, non-randomized building method is in contrast with random forests.  The original random forests procedure suggested by \cite{randomforests} dictates that at each node in each tree, the split may occur on only a randomly selected subset of features.  Thus, we may think of each tree in a random forest as having an additional randomization parameter $\randparam$ that determines the eligible features that may potentially be split at each node.  In a general U-statistic context, we can write this \emph{random kernel} U-statistic as 

\begin{equation}
\label{RFestimator}
U_{\omega; n,k_n,m_n} = \frac{1}{m_{n}} \sum_{(i)} h_{k_n}^{(\randparam_i)} (Z_{i_1}, ..., Z_{i_{k_n}})
\end{equation}

\noindent so that we can write a random forest estimator as 

\[
\rf(\predpoint) = \frac{1}{m_{n}} \sum_{(i)} \treefn_{\predpoint,k_n}^{(\randparam_i)} ((\bm{X}_{i_1},Y_{i_1}), ..., (\bm{X}_{i_{k_n}},Y_{i_{k_n}})).
\]

Due to this additional randomness, random forests and random kernel U-statistics in general do not fit within the framework developed in the previous section so we develop new theory for this expanded class.  Suppose $\randparam_1, ..., \randparam_{m_n} \iid F_{\randparam}$ and that these randomization parameters are selected independently of the original sample $Z_1, ..., Z_n$.  Consider the statistic

\[
U_{\omega; n,k_n,m_n}^{*} = \mathbb{E}_{\randparam} \left( \frac{1}{m_{n}} \sum_{(i)} h_{k_n}^{(\randparam_i)} (Z_{i_1}, ..., Z_{i_{k_n}}) \right)
\]


\noindent so that $U_{\omega; n,k_n,m_n}^{*} = \mathbb{E}_{\omega} U_{\omega; n,k_n,m_n}$.  Taking the expectation with respect to $\omega$, the kernel becomes fixed and hence $U_{\omega; n,k_n,m_n}^{*}$ conforms to the non-random kernel U-statistic theory.  Thus, $U_{\omega; n,k_n,m_n}^{*}$ is asymptotically normal in both the complete and incomplete cases, as well as in the complete and incomplete infinite order cases, by Theorem \ref{subbaggingthm}.  Given this asymptotic normality of $U_{\omega; n,k_n,m_n}^{*}$, in order to retain asymptotic normality of the corresponding random kernel version, we need only show that 

\[
\sqrt{n} (U_{\omega; n,k_n,m_n}^{*} - U_{\omega; n,k_n,m_n}) \inprob 0.
\]

\noindent We make use of this idea in the proof of the following theorem, which is provided in Appendix A.


\begin{theorem}
\label{RFthm}
Let $U_{\omega; n,k_n,m_n}$ be a random kernel U-statistic of the form defined in equation (\ref{RFestimator}) such that $U_{\omega; n,k_n,m_n}^{*}$ satisfies Condition 1 and suppose that $\mathbb{E}h_{k_n}^{2}(Z_1, ..., Z_{k_n}) < \infty$ for all $n$, $\lim \frac{k_n}{\sqrt{n}} = 0$, and $\lim \frac{n}{m_n} = \alpha$.  Then, letting $\beta$ index the subsamples, so long as $\lim \zeta_{1,k_n} \neq 0$ and
\[
\lim_{n \rightarrow \infty} \mathbb{E} \left( h_{k_n}^{(\omega)}(Z_{\beta_{1}}, ..., Z_{\beta_{k_n}}) - \mathbb{E}_{\omega} h_{k_n}^{(\omega)}(Z_{\beta_{1}}, ..., Z_{\beta_{k_n}}) \right)^2 \neq \infty,
\]
\noindent $U_{\omega; n,k_n,m_n}$ is asymptotically normal and the limiting distributions are the same as those provided in Theorem~\ref{subbaggingthm}.
\end{theorem}

Note that the variance parameters $\zeta_{1,k_n}$ and $\zeta_{k_n,k_n}$ in the context of random kernel U-statistics are still defined as the covariance between estimates generated by the (now random) kernels.  Thus, in the specific context of random forests, these variance parameters correspond to the covariance between predictions generated by trees, but each tree is built according to its own randomization parameter $\omega$ and this covariance is taken over $\omega$ as well.  The final condition of Theorem \ref{RFthm} that 


\[
\lim_{n \rightarrow \infty} \mathbb{E} \left( h_{k_n}^{(\omega)}(Z_{\beta_{1}}, ..., Z_{\beta_{k_n}}) - \mathbb{E}_{\omega} h_{k_n}^{(\omega)}(Z_{\beta_{1}}, ..., Z_{\beta_{k_n}}) \right)^2 \neq \infty,
\]

\noindent simply ensures that the randomization parameter $\omega$ does not continually pull predictions from the same subsample further apart as $n \rightarrow \infty$.  This condition is satisfied, for example, if the response $Y$ is bounded and should also be easily satisfied for any reasonable implementation of random forests.

The subsampled random forest algorithm that produces asymptotically normal predictions is provided in Algorithm \ref{algo:rf}.  As with subbagging, this subsampled random forest algorithm is exactly a random forest with subsamples used to build trees instead of full bootstrap samples.

\begin{algorithm}
\begin{algorithmic}
 
\STATE Load training set
\STATE Select size of subsamples $k_n$ and number of subsamples $m_n$

\FOR {$i$ in 1 to $m_n$}

	\STATE Select subsample of size $k_n$ from training set
	\STATE Build tree based on randomization parameter $\omega_i$
	\STATE Use this tree to predict at $\predpoint$
	
\ENDFOR

\STATE Average the $m_n$ predictions to obtain final estimate $\rf(\predpoint)$
\end{algorithmic}
\caption{Subsampled Random Forest}
\label{algo:rf}
\end{algorithm}

Another random-forest-type estimator based on a crossed design that results in an infinite order \emph{generalized} U-statistic is provided in Appendix B.  However, the above formulation most resembles Breiman's original procedure and is more computationally feasible than the method mentioned in Appendix B, so we consider only this random kernel version of random forests in the simulations and other work that follows.


\section{Estimating the Variance}
\label{sec:EstimatingTheVariance}

The limiting distributions provided in Theorem \ref{subbaggingthm} depend on the unknown mean parameter $\theta_{k_n} = \mathbb{E}U_{n,k_n,m_n}$ as well as the unknown variance parameters $\zeta_{1,k_n}$ and $\zeta_{k_n,k_n}$.  In order for us to be able to use these distributions for statistical inference in practice, we must establish consistent estimators of these parameters.  It is obvious that we can use the sample mean -- i.e. the prediction from our ensemble -- as a consistent estimate of $\theta_{k_n}$, but determining an appropriate variance estimate is less straightforward.  

In equation (\ref{ZetaDefn}) of the previous section, we defined $\zeta_{c,k_n}$ as the covariance between two instances of the kernel with $c$ shared arguments, so the sample covariance between predictions may serve as a consistent estimator for both $\zeta_{1,k_n}$ and $\zeta_{k_n,k_n}$.  However, in practice we find that this often results in estimates close to 0, which may then lead to an overall negative variance estimate.

It is not difficult to show - see \cite{Lee1990} page 11 for details - that an equivalent expression for $\zeta_{c,k_n}$ is given by

\[
\zeta_{c,k_n} = var \Big( \mathbb{E} \big( h_{k_n}(Z_1, ..., Z_{k_n})\; | \; Z_1 = z_1, ..., Z_c=z_c \big) \Big).
\]

\noindent To estimate $\zeta_{c,k_n}$ for our tree-based ensembles, we begin by selecting $c$ observations $\tilde{\bm{z}}_1, ..., \tilde{\bm{z}}_c$, which we refer to as initial fixed points, from the training set.  We then select several subsamples of size $k_n$ from the training set, each of which must include $\tilde{\bm{z}}_1, ..., \tilde{\bm{z}}_c$, build a tree with each subsample, and record the mean of the predictions at $\predpoint$.  Let $n_{MC}$ (MC for ``Monte Carlo") denote the number of subsamples drawn so that this average is taken over $n_{MC}$ predictions.  We then repeat the process for $n_{\tilde{\bm{z}}}$ initial sets of fixed points and take our final estimate of $\zeta_{c,k_n}$ as the variance over the $n_{\tilde{\bm{z}}}$ final averages, yielding the estimator

\begin{equation*}
\hat{\zeta}_{c,k_n} = var \Bigg( \frac{1}{n_{MC}} \sum_{i=1}^{n_{MC}} \treefn_{\predpoint,k_n} (\mathcal{S}_{\tilde{\bm{z}}^{(1)},i}) \; , ..., \; \frac{1}{n_{MC}} \sum_{i=1}^{n_{MC}} \treefn_{\predpoint,k_n} (\mathcal{S}_{\tilde{\bm{z}}^{(n_{\tilde{\bm{z}}})},i}) \Bigg)
\end{equation*}

\noindent where $\tilde{\bm{z}}^{(j)}$ denotes the $j^{th}$ set of initial fixed points and $\mathcal{S}_{\tilde{\bm{z}}^{(j)},i}$ denotes the $i^{th}$ subsample that includes $\tilde{\bm{z}}^{(j)}$ (which is used here as shorthand for the argument to the tree function $\treefn_{\predpoint,k_n}$).  Now, since we assume that the orginal data in the training set is i.i.d., the random variables $\frac{1}{n_{MC}} \sum_{i=1}^{n_{MC}} \treefn_{\predpoint,k_n} (\mathcal{S}_{\tilde{\bm{z}}^{(1)},i})$ are also i.i.d.\ and since the sample variance is a U-statistic, $\hat{\zeta}_{c,k_n}$ is a consistent estimator.  The algorithm for calculating $\hat{\zeta}_{1,k_n}$ is provided in Algorithm \ref{algo:zeta1}.  Note that when $c = k_n$, each of the subsamples is identical so we need only use $n_{MC}=1$ which simplifies the estimation procedure for $\zeta_{k_n,k_n}$.  The procedure for calculating $\hat{\zeta}_{k_n,k_n}$ is provided in Algorithm \ref{algo:zetak}.

\begin{algorithm}
\begin{algorithmic}

\FOR {$i$ in 1 to $n_{\tilde{\bm{z}}}$}

\STATE Select initial fixed point $\tilde{\bm{z}}^{(i)}$

\FOR {$j$ in 1 to $n_{MC}$}

	\STATE Select subsample $\mathcal{S}_{\tilde{\bm{z}}^{(i)},j}$ of size $k_n$ from training set that includes $\tilde{\bm{z}}^{(i)}$
	\STATE Build tree using subsample $\mathcal{S}_{\tilde{\bm{z}}^{(i)},j}$
	\STATE Use tree to predict at $\predpoint$
	
\ENDFOR

\STATE Record average of the $n_{MC}$ predictions 

\ENDFOR

\STATE Compute the variance of the $n_{\tilde{\bm{z}}}$ averages
\end{algorithmic}
\caption{$\zeta_{1,k_n}$ Estimation Procedure}
\label{algo:zeta1}
\end{algorithm}


\begin{algorithm}
\begin{algorithmic}

\FOR {$i$ in 1 to $n_{\tilde{\bm{z}}}$}

\STATE Select subsample of size $k_n$ from training set
\STATE Build tree using subsample this subsample
\STATE Use tree to predict at $\predpoint$

\ENDFOR

\STATE Compute the variance of the $n_{\tilde{\bm{z}}}$ predictions
\end{algorithmic}
\caption{$\zeta_{k_n,k_n}$ Estimation Procedure}
\label{algo:zetak}
\end{algorithm}


Choosing the values of $n_{\tilde{\bm{z}}}$ and $n_{MC}$ will depend on the situation.  The number of iterations required to accurately estimate the variance depends on a number of factors, including the tree building method and true underlying regression function.  Of course, ideally these estimation parameters should be chosen as large as is computationally feasible.  In our simulations, we find that in most cases, only a relatively small number of initial fixed point sets are needed, but many more Monte Carlo samples are often necessary for accurate estimation.  In most cases, we used an $n_{MC}$ of at least 500.  Recall that because our trees are built with small subsamples, we can build correspondingly more trees at the same computational cost.

\subsection*{Internal vs External Estimation}

The algorithms for producing the subbagged or subsampled random forest predictions as well as the above algorithms for estimating the variance parameters are all that is needed to perform statistical inference.  We can begin with Algorithm \ref{algo:subbagging} or \ref{algo:rf} to generate the predictions, followed by Algorithms \ref{algo:zeta1} and \ref{algo:zetak} to estimate the variance parameters $\zeta_{1,k_n}$ and $\zeta_{k_n,k_n}$.  This procedure of running these 3 algorithms seperately is what we will refer to as the \emph{external} variance estimation method, since the the variance parameters are estimated outside of the orginal ensemble.  By contrast, we could instead generate the predictions and estimate the variance parameters in one procedure by taking the mean and variance of the predictions generated by the trees used to estimate $\zeta_{1,k_n}$.  Algorithm \ref{algo:internal} outlines the steps in this \emph{internal} variance estimation method.

\begin{algorithm}
\begin{algorithmic}

\FOR {$i$ in 1 to $n_{\tilde{\bm{z}}}$}

\STATE Select initial fixed point $\tilde{\bm{z}}^{(i)}$

\FOR {$j$ in 1 to $n_{MC}$}

	\STATE Select subsample $\mathcal{S}_{\tilde{\bm{z}}^{(i)},j}$ of size $k_n$ from training set that includes $\tilde{\bm{z}}^{(i)}$
	\STATE Build tree using subsample $\mathcal{S}_{\tilde{\bm{z}}^{(i)},j}$
	\STATE Use tree to predict at $\predpoint$ and record prediction
	
\ENDFOR

\STATE Record average of the $n_{MC}$ predictions 

\ENDFOR

\STATE Compute the variance of the $n_{\tilde{\bm{z}}}$ averages to estimate $\zeta_{1,k_n}$
\STATE Compute the variance of all predictions to estimate $\zeta_{k_n,k_n}$
\STATE Compute the mean of all predictions to estimate $\theta_{k_n}$
\end{algorithmic}
\caption{Internal Variance Estimation Method}
\label{algo:internal}
\end{algorithm}

This internal variance estimation method is more computationally efficient and has the added benefit of producing variance estimates by simply changing the way in which the subsamples are selected.  This means that we are able to obtain all parameter estimates we need to conduct inference at no greater computational cost than building the original ensemble.  Although Theorems \ref{subbaggingthm} and \ref{RFthm} dictate that the subsamples used in the ensemble be selected uniformly at random, we find that the additional correlation introduced by selecting the subsamples in this way and using the same subsamples to estimate all parameters does not affect the limiting distribution.
 


\section{Inference Procedures}
\label{sec:InferenceProcedures}

In this section, we describe the inference procedures that may be carried out after performing the estimation procedures.

\subsection{Confidence Intervals}
\label{ConfidenceIntervals}
In Section~\ref{sec:TreesAsUStatistics}, we showed that predictions from subbagging and subsampled random forests are asymptotically normal and in Section~\ref{sec:EstimatingTheVariance} we provided consistent estimators for the parameters in the limiting normal distributions.  Thus, given a training set, we can estimate the approximate distribution of predictions at any given feature vector of interest $\predpoint$.  To produce a confidence interval for predictions at $\predpoint$, we need only estimate the variance parameters and take quantiles from the appropriate limiting distribution.  Formally, our confidence interval is $[LB, UB]$ where the lower and upper bounds, $LB$ and $UB$, are the $\alpha/2$ and $1 - \alpha/2$ quantiles respectively of the normal distribution with mean $\hat{\theta}_{k_n}$ and variance $\frac{k_{n}^{2}}{\hat{\alpha}} \hat{\zeta}_{1,k_n} + \hat{\zeta}_{k_n,k_n}$ where $\hat{\zeta}_{1,k_n}$ and $\hat{\zeta}_{k_n,k_n}$ are the variance estimates and $\hat{\alpha} = n/m_n$.  This limiting distribution is that given in result $(ii)$ of Theorem \ref{subbaggingthm} which is the distribution we recommend using in practice.  

As mentioned in the introduction, these confidence intervals can also be used to address hypotheses of the form

\begin{align*}
&H_0: \; \theta_{k_n} = c \\
&H_1: \; \theta_{k_n} \neq c.
\end{align*}

\noindent Formally, we can define the test statistic

\[
t = \frac{\hat{\theta}_{k_n} - c}{sd(\hat{\theta}_{k_n})}
\]

\noindent and reject $H_0$ if $|t|$ is greater than the the $1-\frac{\alpha}{2}$ quantile of the standard normal.  This corresponds to a test with type 1 error rate $\alpha$ so that $ P[ \text{reject $H_0$} \;|\; \text{$H_0$ true}] = P[\; |t| > 1 - \frac{\alpha}{2} \;|\; \theta_{k_n} = c] = \alpha$.  However, this testing procedure is equivalent to simply checking whether $c$ is within the calculated confidence interval:  if $c$ is in the confidence interval, then we fail to reject this hypothesis that the true mean prediction is equal to $c$, otherwise we reject.


Finally, recall that these confidence intervals are for the expected prediction $\theta_{k_n}$ and not necessarily for the true value of the underlying regression function $\theta = F(\predpoint)$.  If the tree building method employed is consistent so that $\theta_{k_n} \inprob \theta$, then as the sample size increases, the tree should be (on average) producing more accurate predictions, but in order to claim that our confidence intervals are asymptotically valid for $\theta$, we need for this convergence to occur at rate of $\sqrt{n}$ or faster.  However, in general, the rate of convergence will depend on not only the tree-building method and true underlying regression function, but also on the location of the prediction point within the feature space.  

Note that Theorems 1 and 2 apply not only to tree-based ensembles, but to any estimator that can be written in the form of an infinite order U-statistic, as in equation (\ref{TreeEstGrowingRank}).  Some of these ensembles may be straightforward to analyze, but for others, such as random forest predictions near the edge of the feature space, it may be difficult to establish a universal rate of consistency.  However, even when $\sqrt{n}$-consistency cannot be guaranteed, these intervals still provide valuable information not currently available with existing tools.  In these cases, the confidence interval provides a reasonable range of values for where the prediction might fall if the ensemble was recomputed using a new training set; areas of the feature space where confidence intervals are relatively large indicate regions where the ensemble is particularly unstable.  

Compare this, for example, to the standard approach of withholding some (usually small) portion of the training set and comparing predictions made at these hold-out points to the corresponding known responses.  Such an approach provides some information as to the \emph{accuracy} of the learner at specific locations throughout the feature space, but says nothing about the \emph{stability} of these predictions.  Thus, instead of relying only on measures of overall goodness-of-fit such as MSE or SSE, these intervals allow users to investigate prediction variability at particular points or regions and in this sense, can be seen as a measure of how much the accuracy of predictions at that point is due to chance.   


\subsection{Tests Of Significance}
\label{TestsOfSignificance}
The limiting distributions developed in Theorems \ref{subbaggingthm} and \ref{RFthm} also allow us a way to test the significance of features.  In many situations, data are recorded for a large number of features but a sparse true regression structure is suspected.  Suppose that the training set consists of $d$ features, $X_1, ..., X_d$ and consider a reduced set $\xred \subset \{ X_1, ..., X_d \}$.  Let $\predpoints = \{ \bm{x}_1, ..., \bm{x}_N \}$ be a set of feature vectors where we are interested in making predictions.  Also, let $g$ denote the function that maps feature vectors to their corresponding true mean prediction and let $\gred$ denote the same type of function that maps from the reduced feature space.  That is, for a particular prediction point of interest $\predpoint$, $g(\predpoint)$ is the true mean prediction $\theta_{k_n}$ generated by trees built using the full feature space, and $\gred(\predpoint)$ is the true mean prediction $\theta_{k_n}^{(R)}$ generated by trees that are only permitted to utilize features in the reduced set.  Then, for each test point $\featvect \in \predpoints$, we would like to know whether $g(\featvect) = \gred(\featvect)$ so that we can determine the predictive influence of features not in $\xred$.  More formally, we would like to test the hypothesis 

\begin{align}
& H_0:  g(\featvect) = \gred(\featvect) \; \; \forall \; \featvect \in \predpoints \label{HTlab1} \\
& H_1:  g(\featvect) \neq \gred(\featvect) \; \; \text{for some} \; \featvect \in \predpoints. \notag 
\end{align}

\noindent Rejecting this null hypothesis means that a feature not in the reduced feature space $\xred$ makes a significant contribution to the prediction at at least one of the test points.  

To perform this test with a training set of size $n$, we take $m_n$ subsamples, each of size $k_n$, and build a tree with each subsample.  Denote these subsamples $S_1, ..., S_{m_n}$ and for a given feature vector $\featvect$, let $\hat{g}(\featvect)$ denote the average over the predictions at $\featvect$ generated from the $m_n$ trees.  Then, using the same subsamples $S_1, ..., S_{m_n}$, again build a tree with each, but using only those features in $\xred$, and let $\gredhat(\featvect)$ be the average prediction at $\featvect$ generated by these trees.  Finally, define the difference function

\[
\hat{D}(\featvect) = \hat{g}(\featvect) - \gredhat(\featvect)
\]

\noindent as the difference between the two ensemble predictions.  Note that we can write

\begin{align*}
\hat{D}(\featvect) &= \hat{g}(\featvect) - \gredhat(\featvect) \\
&= \frac{1}{m_n} \sum_{(j)} \treefn_{\featvect,k_n}(S_j) - \frac{1}{m_n} \sum_{(j)} \treefn_{\featvect,k_n}^{(R)} (S_j) \\
&= \frac{1}{m_n} \sum_{(j)} \Big( \treefn_{\featvect,k_n}(S_j) - \treefn_{\featvect,k_n}^{(R)}(S_j) \Big)
\end{align*}

\noindent so that this difference function is a U-statistic.  Thus, if we have only a single test point of interest, $\hat{D}$ is asymptotically normal, so $\hat{D}^2$ is asymptotically $\chi_{1}^{2}$ and we can use $\hat{D}^2$ as a test statistic.  

However, it is more often the case that we have several test points of interest.  In this case, define $\hat{\mathbb{D}}$ to be the vector of observed differences in the predictions

\[
\hat{\mathbb{D}} = \big( \hat{D}(\bm{x}_1), ..., \hat{D}(\bm{x}_N) \big)
\]

\noindent so that, provided a joint distribution exists with respect to Lebesgue measure, $\hat{\mathbb{D}}$ has a multivariate normal distribution with mean vector

\[
\bm{\mu} = \left( g(\bm{x}_1) - \gred(\bm{x}_1), ...,g(\bm{x}_N) - \gred(\bm{x}_N) \right)^{T}
\]

\noindent which we estimate with 

\[
\hat{\bm{\mu}} = \hat{\mathbb{D}}^{T}
\]

\noindent as well as a covariance matrix $\Sigma$.  This covariance matrix has parameters $\Sigma_{1,k_n}$ and $\Sigma_{k_n,k_n}$, the multivariate analogues of $\zeta_{1,k_n}$ and $\zeta_{k_n,k_n}$.  Consistent estimators for these multivariate parameters can be obtained by simply replacing the variance calculation in Algorithms \ref{algo:zeta1} and \ref{algo:zetak} with a covariance.  For clarity, the procedure for obtaining $\hat{\Sigma}_{1,k_n}$ is provided in Algorithm \ref{algo:Sigma1Est} in Appendix C.

Finally, combining these predictions to form a consistent estimator $\hat{\Sigma}$ we have that 

\[
\hat{\bm{\mu}}^T \hat{\Sigma}^{-1} \hat{\bm{\mu}} \sim \chi_{N}^2
\]

\noindent under $H_0$.  Thus, in order to test the hypothesis in (\ref{HTlab1}), we compare the test statistic $\hat{\bm{\mu}}^T \hat{\Sigma}^{-1} \hat{\bm{\mu}}$ to the $1 - \alpha$ quantile of the $\chi_{N}^{2}$ distribution to produce a test with type 1 error rate $\alpha$.  If our test statistic is larger than this critical value, we reject the null hypothesis.


\subsection*{Further Testing Procedures}

This setup, though straightforward, may not always definitively decide the significance of features.  In some cases, even randomly generated features that are unrelated to the response can be reported significant.  Depending on the building method, tree-based algorithms may take advantage of additional randomness in features even when the particular values of those features do not directly contribute to the response.  For this reason, we also recommend repeating the testing procedure by comparing predictions generated using the full dataset to predictions generated by a dataset with randomly generated values -- commonly obtained by permuting the values in the training set -- for the features not in the reduced feature set to test hypostheses of the form

\begin{align*}
\label{testHyp} & H_0:  g(\featvect) = g^{(RAND)}(\featvect) \; \; \forall \; \featvect \in \predpoints \\
& H_1:  g(\featvect) \neq g^{(RAND)}(\featvect) \; \; \text{for some} \; \featvect \in \predpoints. \notag 
\end{align*}

The testing procedure remains exactly the same except that to calculate the second set of trees,  we simply substitute the reduced training set for a training set with the same number of features, but with randomized values taking the place of the original values for the additional features.  Rejecting this null hypothesis allows us to conclude that not only do the additional features not in the reduced training set make a significant contribution to the predictions, but that the contribution is significantly more than could be obtained simply by adding additional randomness.


There are also two additional tests that may be performed.  First, we can test whether predictions generated by a training set with randomized values for the additional features are significantly different from predictions generated by the reduced feature set.  If a significant difference is found, then the trees in the ensemble are making use of the additional randomness or possibly an accidental structure in the randomized features.  As a final check, we can compare predictions generated by two training sets, each with randomized values for the features not in the reduced set.  In the unlikely event that a significant difference is found between these predictions, it is again likely due to an accidental structure in the randomized values.  Both of these tests can be performed in exactly the same fashion by substituting the appropriate training sets.

\section{Simulations}
\label{sec:Simulations}

We present here a small simulation study in order to illustrate the limiting distributions derived in Section \ref{sec:TreesAsUStatistics} and also to demonstrate the inference procedures proposed in the previous section.  We consider two different underlying regression functions:

\begin{enumerate}
\item $g(x_1) = 2x_1$; \hspace{3mm} $\mathcal{X} = [0,20]$
\item $g(\bm{x}) = 10 \sin(\pi x_1 x_2) + 20(x_3 - 0.05)^2 + 10 x_4 + 5 x_5$; \hspace{3mm} $\mathcal{X} = [0,1]^5$
\end{enumerate}

\noindent The first function corresponds to simple linear regression (SLR) and was chosen for simplicity and ease of visualization.  The second was initially considered by \cite{Friedman1991} in development of the Multivariate Adaptive Regression Spline (MARS) procedure and was recently investigated by \cite{biau12}.  In each case, features were selected uniformly at random from the feature spaces and responses were sampled from $g(\bm{x}) + \epsilon$, where $\epsilon \iid \mathcal{N}(0,10)$, to form the training sets.


\subsection*{Limiting Distributions}

\begin{figure}
  \centering
  \includegraphics[scale=0.32]{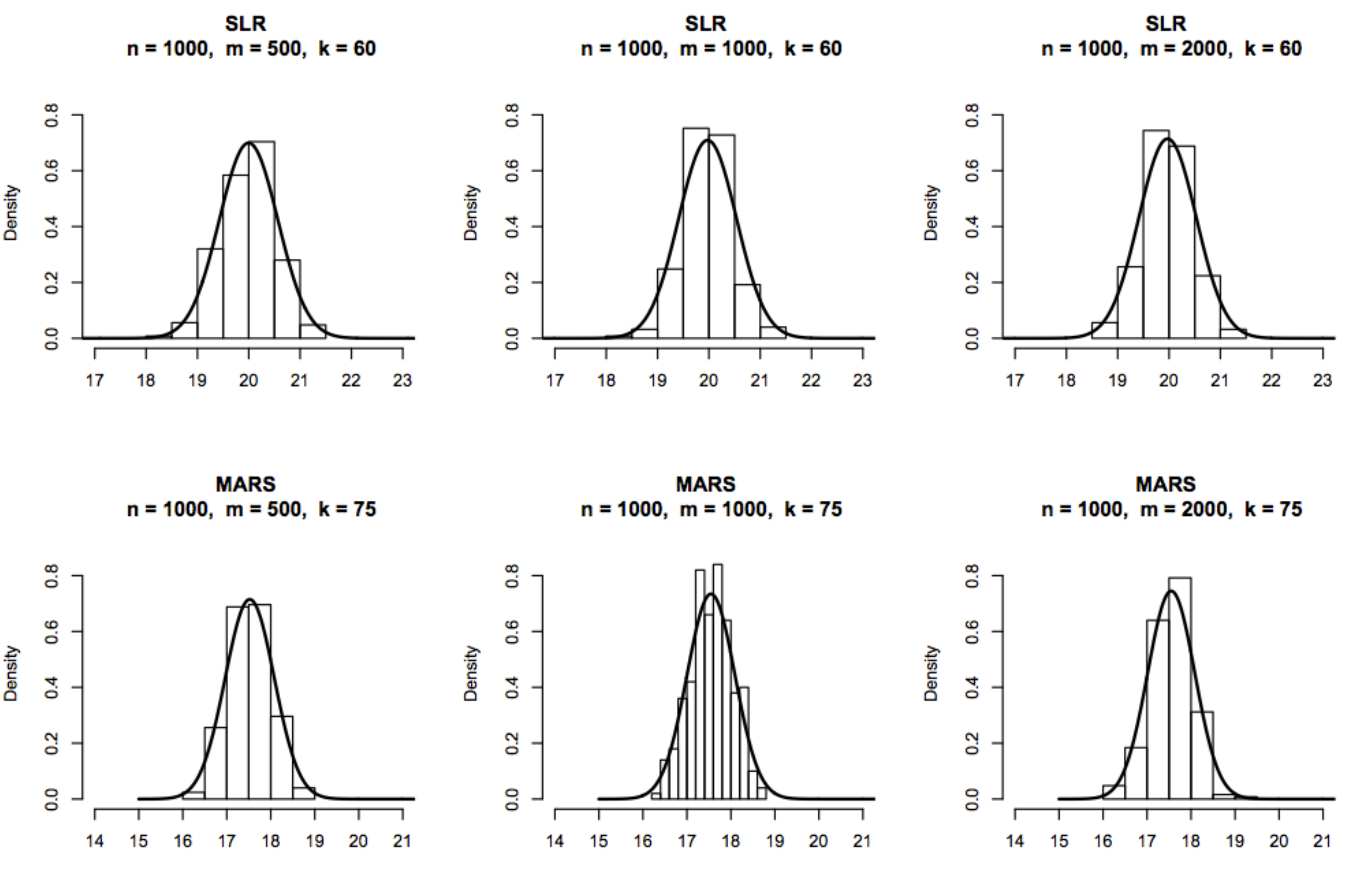}
  \caption{\label{fig:SLRandMARShist}  Histograms of subbagged predictions at $x_1 = 10$ in the SLR case (top row) and at $x_1= \cdots = x_5 = 0.5$ in the MARS case (bottom row).  The total sample size, number of subsamples, and size of each subsample are denoted by $n$,$m$, and $k$, respectively in the plot titles.}
\end{figure}

We begin by illustrating the distributions of subbagged predictions.  In the SLR case, predictions were made at $x_1 = 10$ and in the MARS case, predictions were made at $x_1= \cdots = x_5 = 0.5$.  The histograms of subbagged predictions are shown in Figure \ref{fig:SLRandMARShist}.  Each histogram is comprised of 250 simulations.

For each histogram, the size of the training set $n$, number of subsamples $m$, and size of each subsample $k$, is provided in the title.  Each tree in the ensembles was built using the \verb!rpart! function in \verb!R!, with the additional restriction that at least 3 observations per node were needed in order for the algorithm to consider splitting on that node.  Overlaying each histogram is the density obtained by estimating the parameters in the limiting distribution.  In each case, we take the limiting distribution to be that given in result \emph{(ii)} of Theorem \ref{subbaggingthm}; namely that the predictions are normally distributed with mean $\mathbb{E}b_{n,k,m}(\predpoint)$ and variance $\frac{1}{\alpha}\frac{k^2}{m} \zeta_{1,k} + \frac{1}{m} \zeta_{k,k}$.  

The mean $\mathbb{E}b_{n,k,m}(\predpoint) = \theta_{k}$ was estimated as the empirical mean across the 250 subbagged predictions.  To estimate $\zeta_{k,k}$, 5000 new subsamples of size $k$ were selected and with each subsample, a tree was built and used to predict at $x_1=10$ and $\hat{\zeta}_{k,k}$  was taken as the empirical variance between these predictions.  To estimate $\zeta_{1,k}$, we follow the procedure in Algorithm \ref{algo:zeta1} with $n_{\tilde{\bm{z}}} = 50$ and $n_{MC} = 1000$ in the SLR cases and with $n_{\tilde{\bm{z}}} = 250$ and $n_{MC} = 1000$ in the MARS cases.  Note that since we are only interested in verifying the distributions of predictions, the variance parameters are estimated only once for each case and not for each ensemble.  

It is worth noting that the same variance estimation procedure with $n_{\tilde{\bm{z}}} = 250$ and $n_{MC} = 250$ lead to an overestimate of the variance, so we reiterate that using a large $\nmc$ seems to provide better results, even when $\nxone$ is relatively small.  In each case, we use $\frac{n}{m}$ as a plug-in estimate for $\alpha = \lim \frac{n}{m}$.  We also repeated this procedure and generated the distribution of predictions according to the internal variance estimation method described in Algorithm \ref{algo:internal}.  Details and histograms are provided in Appendix D.  These distributions appear to be the same as when the subsamples are selected uniformly at random, as in the external variance estimation method.

%

\begin{figure}
  \centering
  \includegraphics[scale=0.38]{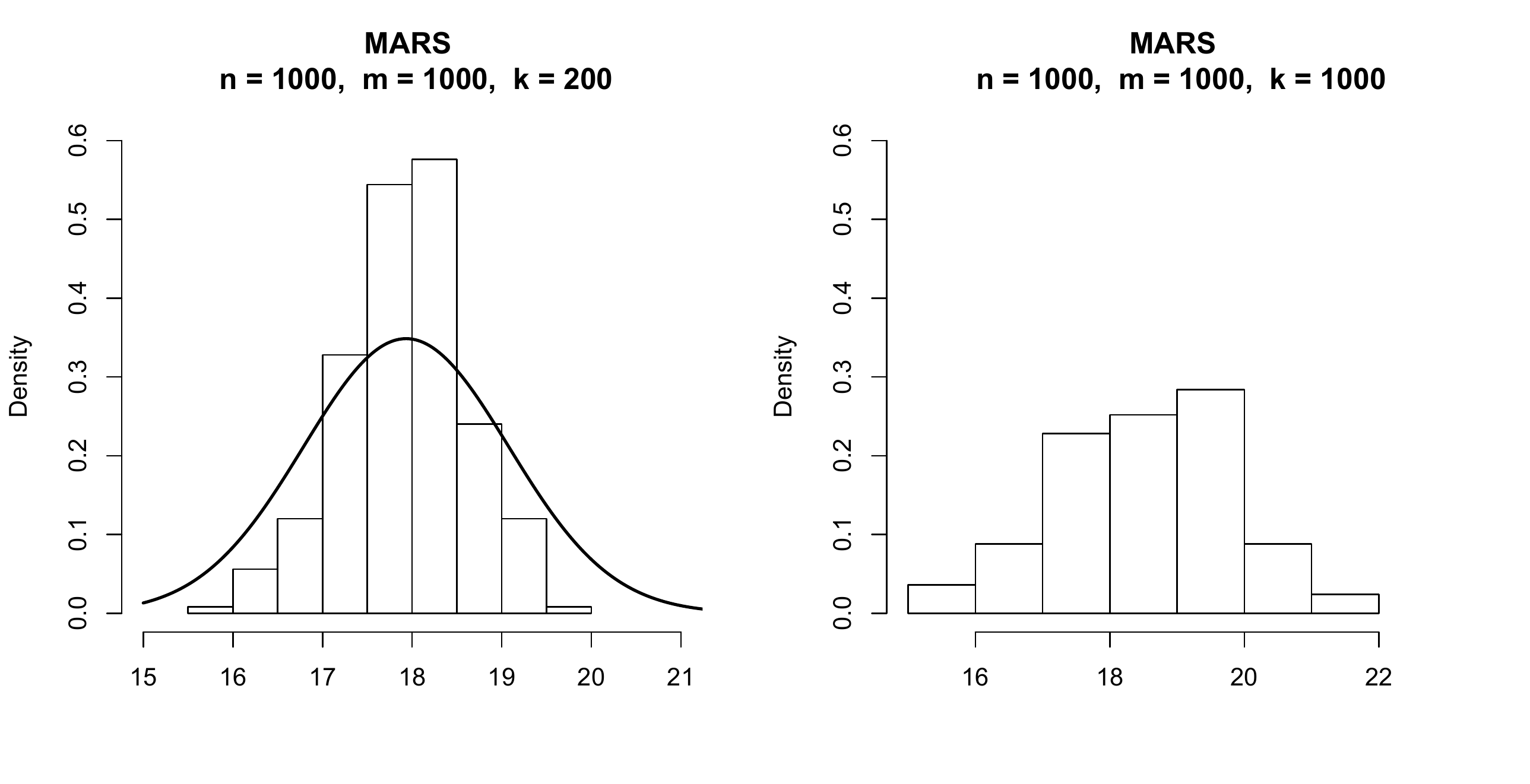}
  \caption{\label{fig:Bigkhist} Histograms of subbagged predictions with larger subsample size $k$ and full bootstrap samples.  Predictions are made at $x_1= \cdots = x_5 = 0.5$.}
\end{figure}

Note that the distributional results in Theorem \ref{subbaggingthm} require $\lim \frac{k}{\sqrt{n}} = 0$, so in practice, the subsample size $k$ should be small relative to $n$.  In the above simulations, we choose $k$ slightly larger than $\sqrt{n}$ and the distributions appear normally distributed with the correct limiting distribution.  However, though this restriction on the growth rate of the subsample size is sufficient for asymptotic normality, it is perhaps not necessary.  In our simulations, we found that ensembles built with larger $k$ are still approximately normal, but begin to look increasingly further from normal as $k$ increases.  The histograms in Figure \ref{fig:Bigkhist} show the distribution of subbagged predictions in the MARS case with $n = m = 1000$ and $k=200$ and also with $n = m = k = 1000$ so that we are using full bootstrap samples to build the ensembles in the latter case.  The parameters in the limiting distribution are estimated in exactly the same manner as with the smaller $k$ for the case where $k = 200$.  In the bootstrap case, we cannot follow our subbagging procedure exactly since the bootstrap samples used to build each tree in the ensemble must be taken with replacement, so we do not attempt to estimate the variance.  These distributions look less normal and we begin to overestimate the variance in the case where $k = 200$.


\subsection*{Confidence Intervals}

We move now to building confidence intervals for predictions and examine their coverage probabilities.  We begin with the SLR case, with $n=200$, $m=200$, and $k=30$ and as above, predict at $x_1=10$.  To build the confidence intervals, we generate 250 datasets and with each dataset, we produce a subbagged ensemble, estimate the parameters in the limiting distribution, and take the 0.025 and 0.975 quantiles from the estimated limiting normal distribution to form an approximate 95\% confidence interval.  The mean of this limiting normal $\theta_{k}$ was estimated as the mean of the predictions generated by the ensemble.  The variance parameter $\zeta_{k,k}$ was estimated by drawing 500 new subsamples, not necessarily used in the ensemble, and calculating the variance between predictions generated by the resulting 500 trees and $\zeta_{1,k}$ was estimated externally using $\nxone = 50$ and $\nmc = 250$.  

\begin{figure}
  \centering
  \includegraphics[scale=0.45]{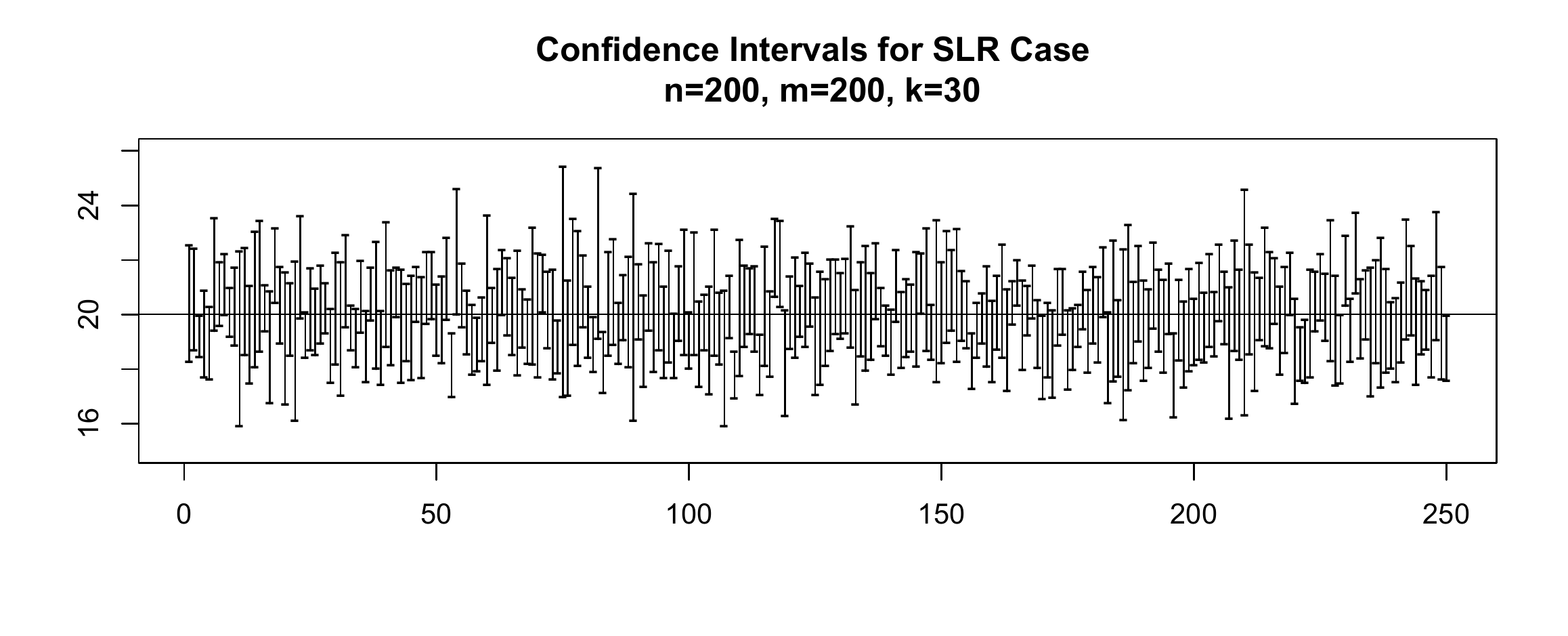}
  \caption{Confidence Intervals for subbagged predictions.}
  \label{fig:SLRCI}
\end{figure}

In order to assess the coverage probability of our confidence intervals, we first need to estimate the true mean prediction $\theta_{k}$ at $x_1=10$ that would be generated by this subbagging ensemble.  To estimate this true mean, we built 1000 subbagged ensembles and took the mean prediction generated by these ensembles, which we found to be 20.02 - very close to the true underlying value of 20.  In this case, we found a coverage probability of 0.912, which means that 228 of our 250 confidence intervals contained our estimate of the true mean prediction.  These confidence intervals are shown in Figure \ref{fig:SLRCI}.  The horizontal line in the plot is at 20.02 and represents our estimate of the true expected prediction.


This same procedure was repeated for the SLR case with $n = m = 1000$ and $k = 60$, the MARS case with $n = m = 500$ and $k = 50$, and the MARS case with $n = m = 1000$ and $k = 75$.  In each case, we produced 250 confidence intervals and the coverage probabilities are shown in Table \ref{TableCovProb}.  The parameters in the limiting distributions were estimated externally in exactly the same fashion, using $\nxone = 50$ and $\nmc = 250$ to estimate $\zeta_{1,k}$.  These slightly higher coverage probabilities mean that we are overestimating the variance, which is likely due to smaller values of the estimation parameters $\nxone$ and $\nmc$ being used to estimate $\zeta_{1,k}$.

\vspace{5mm}
\begin{table}[h]
\begin{tabular}{|l|c|c|c|c|c|l}
\cline{1-6}
\multicolumn{1}{|c|}{\multirow{2}{*}{\begin{tabular}[c]{@{}c@{}}Underlying \\ Function\end{tabular}}} & \multirow{2}{*}{$n$}    & \multirow{2}{*}{$k$}    & \multirow{2}{*}{$\theta_{k}$} & \multicolumn{2}{|c|}{\textbf{Coverage Probability}} &  \\ \cline{5-6}
                                                                                                      & \multicolumn{1}{|l}{} & \multicolumn{1}{|l}{} & \multicolumn{1}{|l|}{}  & External Variance Estimate    & Internal Variance Estimate    &  \\ \cline{1-6}
SLR                                                                                                   & 200                   & 30                    & 19.94                   & 0.912                   & 0.912                     &  \\ \cline{1-6}
SLR                                                                                                   & 1000                  & 60                    & 19.99                   & 0.956                   & 0.936                     &  \\ \cline{1-6}
MARS                                                                                                  & 500                   & 50                    & 17.43                   & 0.980                   & 0.984                     &  \\ \cline{1-6}
MARS                                                                                                  & 1000                  & 75                    & 17.56                   & 0.996                   & 0.996                     &  \\ \cline{1-6}
\end{tabular}
\caption{\label{TableCovProb} Coverage probabilities}
\end{table}
\vspace{5mm}

We also repeated this procedure for generating confidence intervals using the internal variance estimation method.  The resulting coverage probabilities are remarkably similar to the external variance estimation method and are shown in Table \ref{TableCovProb}.  These ensembles were built using $\nxone = 50$ and $\nmc = 250$.


\subsection*{Hypothesis Testing}

We now explore the hypothesis testing procedure for assessing feature significance.  We focus on the MARS case, where our training set now consists of 6 features $X_1, ..., X_6$, but the response $Y$ depends only on the first 5.  The values of the additional feature $X_6$ are sampled uniformly at random from the interval $[0,1]$ and independently of the first 5 features.

\begin{figure}
  \centering
  \includegraphics[scale=0.45]{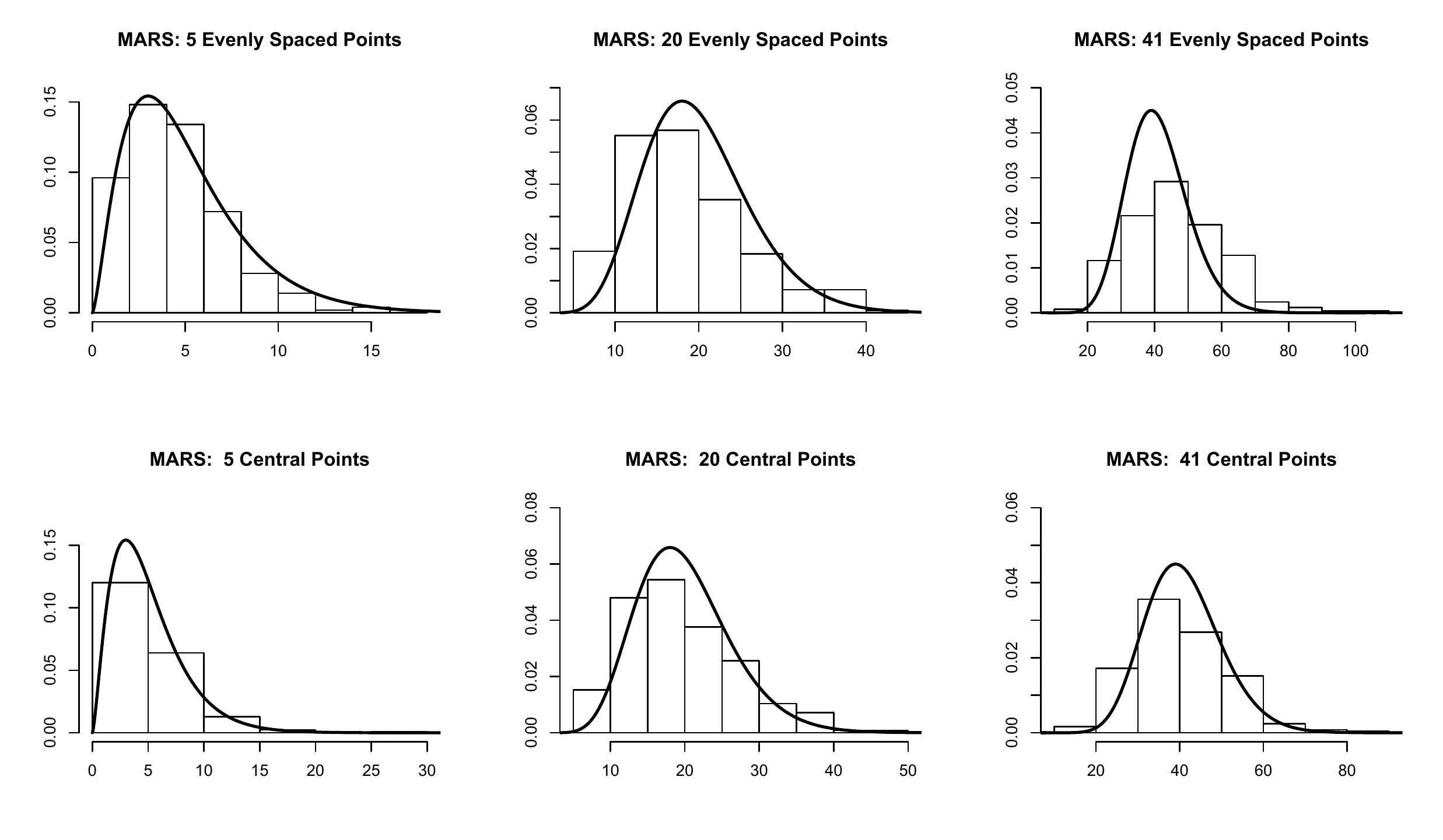}
  \caption{Histograms of simulated test statistics with estimated $\chi^2$ overlay.  The top row of histograms involve test points equally spaced between 0 and 1 and the bottom row corresponds to points randomly selected from the interior of the feature space.}
  \label{HypTestHist}
\end{figure}

We begin by looking at the distribution of test statistics when the test set consists 41 equally spaced points between 0 and 1.  That is, the first test point is $x_1 = \cdots = x_6 = 0$, the second is $x_1 = \cdots = x_6 = 0.025$, and so on so that the last test point is $x_1 = \cdots = x_6 = 1$.  For this test, we are interested in looking at the difference between trees built using all features and those built using only the first 5 so that in the notation in Section \ref{TestsOfSignificance}, $\xred = \{X_1, ..., X_5\}$.  We ran 250 simulations with $n=1000$, $m=1000$, and $k=75$ using a test set consisting of all 41 test points, the 20 central-most points, and the 5 central-most points.  The parameter $\Sigma_{1,k}$ was estimated externally using $\nxone = 100$ and $\nmc = 5000$ and $\Sigma_{k,k}$ was estimated by taking the covariance of the difference in predictions generated by 5000 trees.  These covariance parameters are estimated only once instead of within each ensemble since we are only interested in the distribution of test statistics.  Histograms of the resulting test statistics along with an overlay of the estimated $\chi^2$ densities are shown in the top row of Figure \ref{HypTestHist}. 

We repeated this procedure, this time with a test set consisting of points in the center of the hypercube so we are not predicting close to any edge of the feature space.  The value of each feature in the test set was selected uniformly at random from $[0.25, 0.75]$.  A total of 41 such test points were generated, and histograms of the 250 resulting test statistics were produced in the cases where we use 5 of these test points, 20 test points, and all 41 test points.  These histograms and estimated $\chi^2$ densities are shown in the bottom row of Figure \ref{HypTestHist}.  Note that the bottom row appears to be a better fit and thus there appears to be some bias occuring when test points are selected near the edges of the feature space.  

To check the alpha level of the test -- the probability of incorrectly rejecting the null hypothesis -- we simulated 250 new training sets and used the test set consisting of 41 randomly selected central points.  For each training set, we built full and reduced subbagged estimates, allowed and not allowed to utilize $X_6$ respectively, estimated the parameters, and performed the hypothesis test.  For each ensemble, the variance parameter $\Sigma_{1,k}$ was estimated externally using $\nxone = 50$ and $\nmc = 1000$ and $\Sigma_{k,k}$ was estimated externally on an independently generated set of trees.  

\begin{figure}
  \centering
  \includegraphics[scale=0.37]{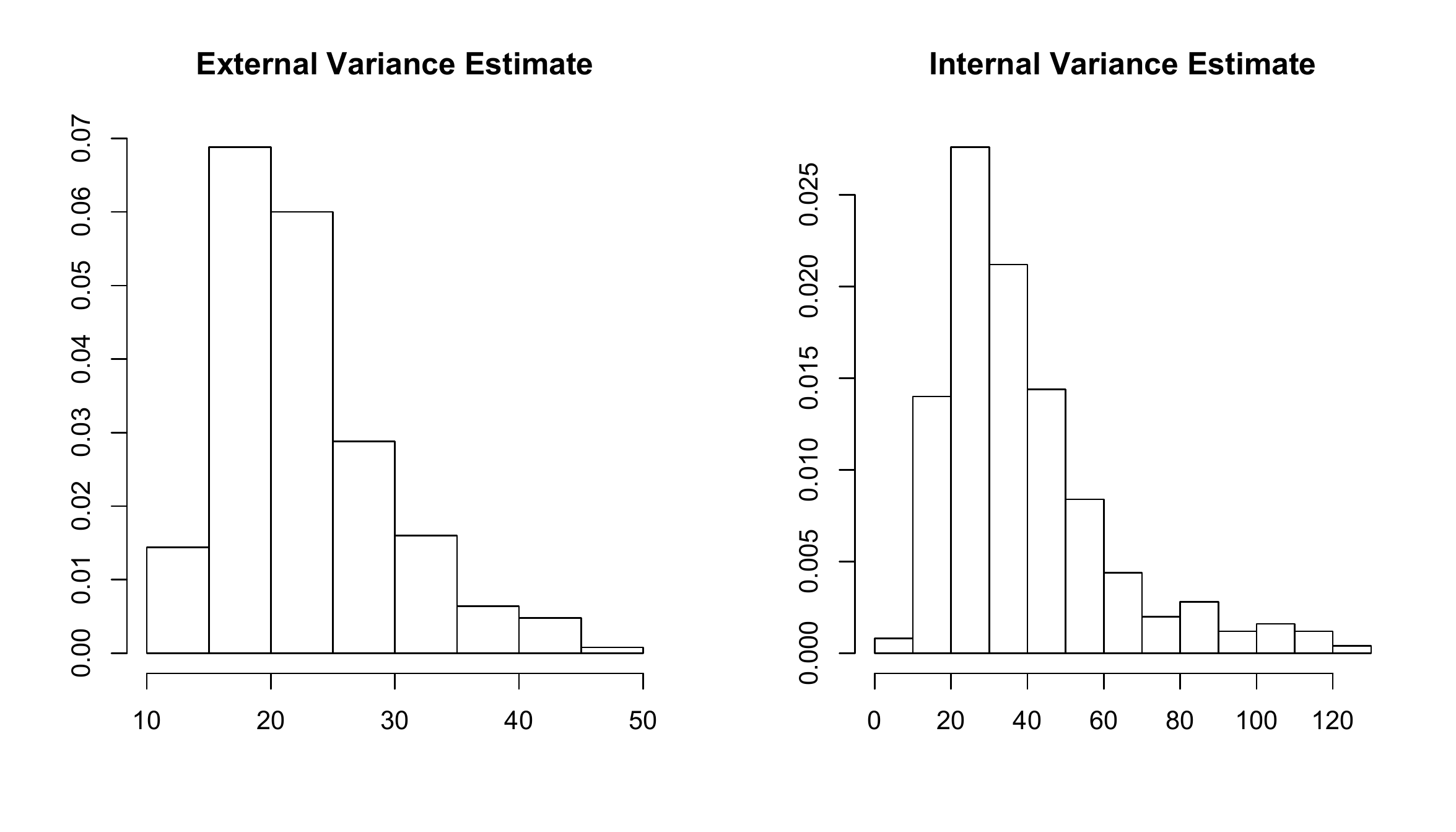}
  \caption{Histograms of simulated test statistics}
  \label{AlphaLevelHist}
\end{figure}

In this setup, none of the 250 simulations resulted in rejecting the null hypothesis, so our empirical alpha level was 0.  A histogram of the resulting test statistics is shown on the left of Figure \ref{AlphaLevelHist}; the critical value, the 0.95 quantile of the $\chi_{41}^{2}$, is 56.942.  Thus, as with the confidence intervals, we are being conservative.  Recall that our confidence interval simulations with $n=1000$, $m=1000$, and $k=75$ predicting at $\bm{x} = (0.5,..., 0.5)$ captured our true value 99.6\% of the time, so this estimate of 0, though conservative, is not necessarily unexpected.  We also repeated this procedure using an internal variance estimate with $\nxone = 50$ and $\nmc = 1000$ and found an alpha level of 0.14.  The histogram of test statistics resulting from the internal variance estimation method is shown on the right in Figure \ref{AlphaLevelHist}.  Here we see that the correlation introduced by not taking an i.i.d. selection of subsamples to build the ensemble may be slightly inflating the test statistics.  

\subsection*{Random Forests}

\begin{figure}
  \centering
  \includegraphics[scale=0.43]{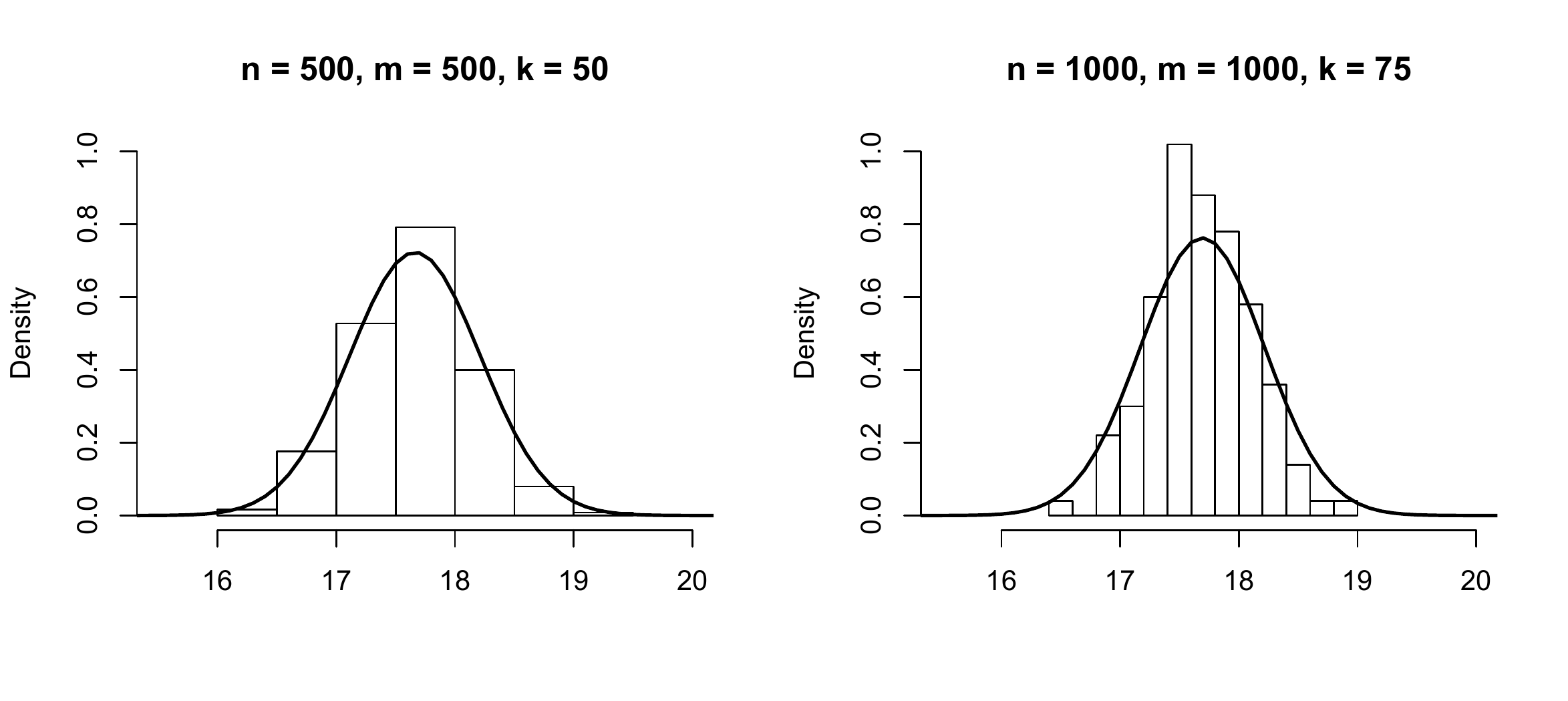}
  \caption{  \label{RFHist} Histograms of random forest predictions at $x_1 = \cdots = x_5 = 0.5$ with estimated normal density overlaid.}
\end{figure}

Thus far, our simulations have dealt only with subbagged ensembles, but recall that Theorem \ref{RFthm} established the asymptotic normality for predictions generated by random forests as well.  The histograms in Figure~\ref{RFHist} show the distribution of predictions generated by subsampled random forests at $x_1 = \cdots = x_5 = 0.5$ when the true underlying function is the MARS function.  These trees were grown using the \verb!randomForest! function in \verb!R! with the \verb!ntree! argument set to 1.  At each node in each tree, 3 of the 5 features $X_1, ..., X_5$ were selected at random as candidates for splits and we insisted on at least 2 observations in each terminal node.  The histograms show the empirical distribution of 250 subsampled random forest predictions and the overlaid density is the limiting normal $\mathcal{N}(\mathbb{E}r_{n,k,m}(\predpoint),\frac{1}{\alpha}\frac{k^{2}}{m} \zeta_{1,k} + \frac{1}{m} \zeta_{k,k})$ with the variance parameters estimated externally.  Our estimate of the mean of this distribution was taken as the empirical mean of the 250 predictions.  Our estimate of $\zeta_{k,k}$ was taken as the empirical variance of predictions across 5000 new trees and the estimate for $\zeta_{1,k}$ was calculated with $\nxone = 250$ and $\nmc = 2000$.

\section{Real Data}
\label{sec:RealData}

We now apply our inference methods to a real dataset provided by the Lab of Ornithology at Cornell University.  The data is part of the ongoing \emph{eBird} citizen science project described in \cite{ebird}.  This project is hosted by Cornell's Lab of Ornithology and relies on citizens, referred to as \emph{birders}, to submit reports of bird observations.  Location, bird species observed and not observed, effort level, and number of birds of each species observed are just a few of the variables participants are asked to provide.  In addition to the data contained in these reports, landcover characteristics as reported in the 2006 United States National Land Cover Database are also available so that information about the local terrain may be used to help predict species abundance.

For our analysis, we restrict our attention to observations (and non-observations) of the Indigo Bunting species.  For the first part of our analysis, we further restrict our attention to observations made during the year 2010.  A little more than 400,000 reports of either presence or absence of Indigo Buntings were recorded during 2010 and the dataset consists of 23 features.  Like many species, the abundance of Indigo Buntings is known to fluctuate throughout the year, so we have two primary goals:  (1) to produce confidence intervals for monthly abundance and  (2) to show that the feature `month' is significant for predicting abundance.

\begin{figure}
  \centering
  \label{fig:IBplot}
  \includegraphics[scale=0.5]{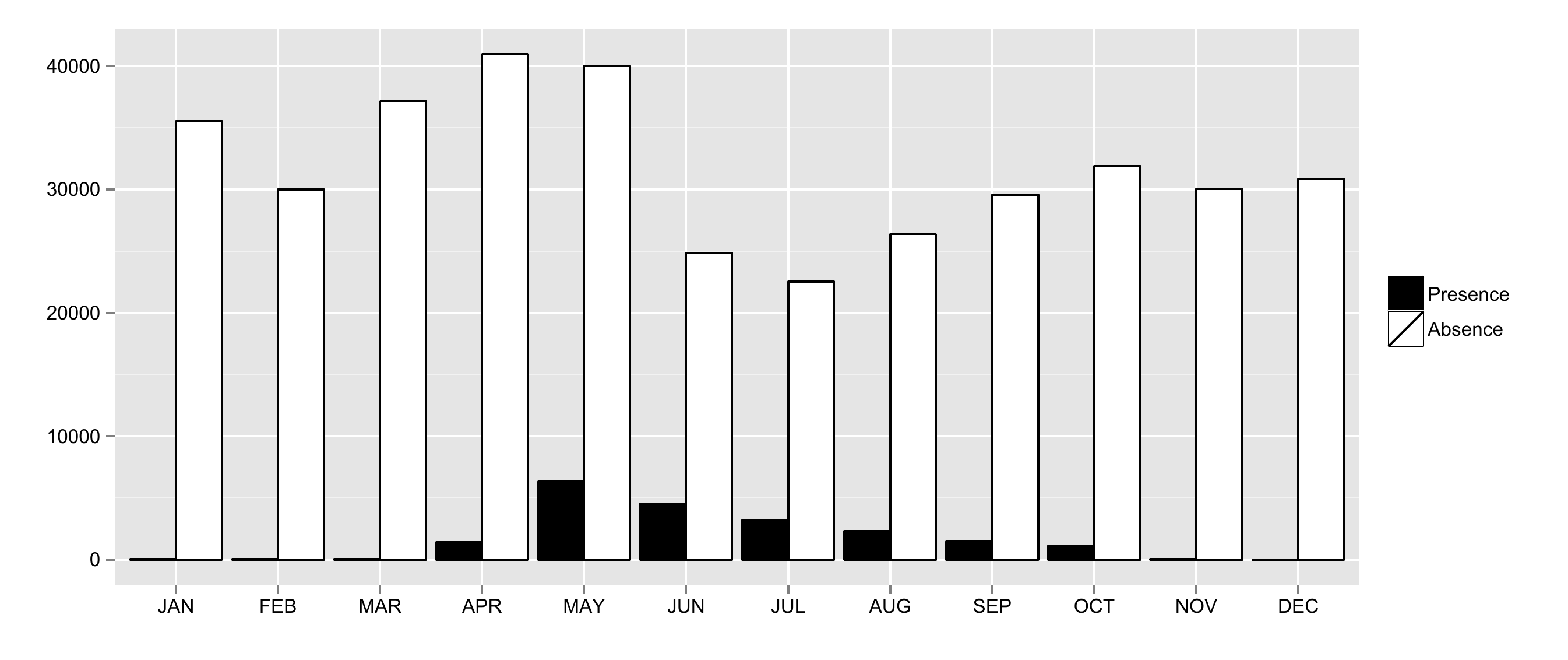}
  \caption{  \label{fig:IBplot} Monthly counts of Indigo Bunting observations.}
\end{figure}

A presence/absence plot of Indigo Buntings by month is shown in Figure \ref{fig:IBplot}.  A few features of this plot are worth pointing out.  Most obviously, there are many more absence observations each month than presence observations.  This makes sense because each time a birder submits a report, they note when Indigo Buntings are not present.  Next, we see that this species is only observed during the warmer months, so month seems highly significant for predicting abundance.  Finally, we see that all months have a large number of reports, so we need not worry about underreporting issues throughout the year.

First we produce the confidence intervals.  The goal is to get an idea of the monthly abundance, which we can think of as `probability of observation', so our test points will be 12 vectors, one for each month.  For the values of the other features, we will use the average of the values recorded for that feature, or, in the case of categorical features, use the most popular category.  Some features, such as elevation, have missing values which are not included in calculating the averages.  We also removed the \emph{day} feature from the training set, since day of the year is able to capture any effect of month.  Since the training set consists of approximately 400,000 observations, we use a subsample size $k = 650$, slightly more than the square root of the training set size.  We build a total of $m = 5000$ trees and take the variance of these predictions at each point to be our estimates of $\zeta_{k,k}$.  We use an external estimate with $\nxone = 250$ and $\nmc = 5000$ to estimate $\zeta_{1,k}$.  For each tree built, we require a minimum of 20 observations to consider splitting an internal node.  We also repeat this procedure using an internal estimate of variance with $\nxone = 250$ and $\nmc = 5000$.

\begin{figure}
  \centering
  \label{fig:ebirdCI}
  \includegraphics[scale=0.5]{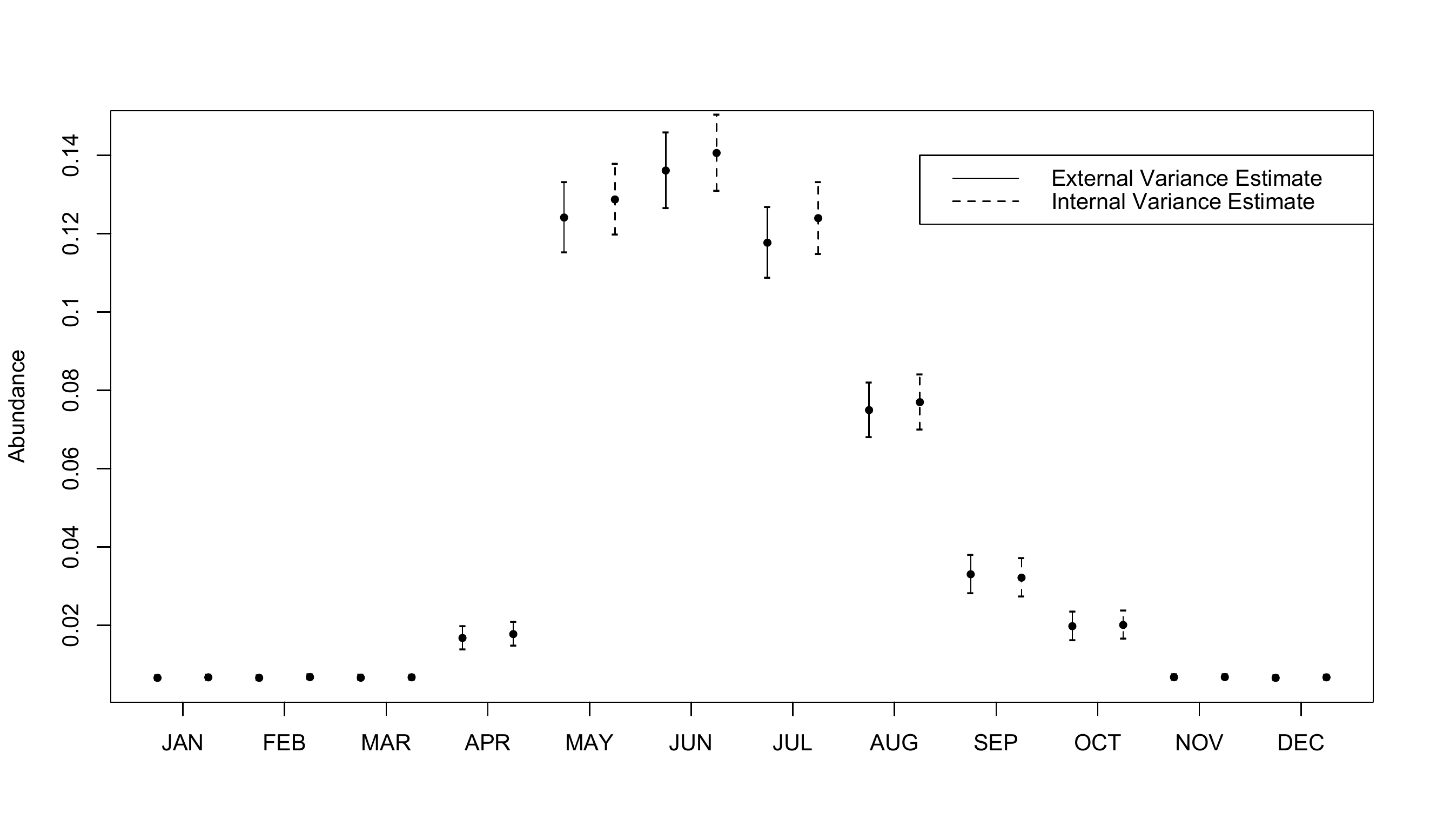}
  \caption{  \label{fig:ebirdCI} Monthly confidence intervals for Indigo Bunting abundance.}
\end{figure}

The confidence intervals are shown in Figure \ref{fig:ebirdCI}.  Note that the pattern seen in Figure \ref{fig:IBplot} is mirrored in the confidence interval plot:  the confidence intervals are higher during months where more positive observations are recorded.  It is also interesting to note that the width of the confidence intervals is larger during months of higher abundance.  Observe in Figure \ref{fig:IBplot} that even for months with many positive observations reported, many more negative observations are also reported.  Thus, for these months there are likely a number of trees in the ensemble with nearly all positive or negative observations so we expect a higher variance in predictions.  For months when very few positive observations are made, nearly all observations in the terminal node will be abscence observations, thus resulting in a very small variance and much narrower confidence intervals.  Based on a visual inspection of the confidence intervals, there appears to be a clear significant difference in abundance between certain months, but we need to account for correlations in our predictions so we also conduct hypothesis tests.  

We conduct these formal tests for the significance of month, following the procedure in Section \ref{TestsOfSignificance}.  To perform this test, we randomly selected 20 points from the training set as the test set and calculated the test statistic based on an internal variance estimate with $\nxone = 250$ and $\nmc = 5000$.  We calculated a test statistic of 4233.10 and a critical value, the 0.95 quantile of the $\chi^{2}_{20}$, of only 31.41 which yields a p-value of approximately 0, so it seems that month is highly significant for predicting abundance.  However, when we generated random values for month and repeated the testing procedure, we calculated a test statistic of 58.02 which, though significantly smaller than the test statistic calculated on the original training set, is still significant.  To ensure that the randomized months we selected did not add any accidental structure, we compared predictions generated using this training set to those generated by another training set, also with randomized months.  Here we find a test statistic of only 2.36 and thus there is no significant difference between these predictions, so the trees are simply taking advantage of additional randomness.  Finally, we test for a difference in predictions generated by the original training set and those generated by the training set with random values of month.  In this case, we calculated a test statistic of 2336.14 which is highly significant so we can conclude that month is significant for predicting abundance and the contribution to the prediction is significantly more than would be expected by simply adding a random feature to the model.

This significant effect of month comes as no surprise as Indigo Buntings are known to be a migratory species.  However, many scientists also believe that migrations may change from year to year and thus year may also be significant for predicting abundance.  For this test, we used the full training dataset consisting of approximately 1 million observations from 2004 to 2010 and as a test set, randomly selected 20 observations from the training set.  Given this larger training set, we increased our subsample size to $k = 1000$ and our Monte Carlo sample size to $n_{MC} = 8000$ and again performed the tests using the internel variance estimation method.  In the first setup where we test predictions from the full training set against predictions generated from the training set without year, we find a test statistic of 94.43 which means that year is significant for making predictions as it is larger than the critical value of 31.41.  However, as was the case in testing the significance of month, we find that a randomly generated year feature is also significant with a test statistic of 52.51.  Following in the same manner as above, we compare predictions from two training sets, each with a randomized year feature, and we find no significant difference in these predictions with a test statistic of only 4.70.  Finally, we test for a difference in predictions between the full training set and a dataset with random year and find that there is a significant difference in these predictions with a test statistic of 109.72.  Thus, as was the case with month, we can conclude that year is significant for predicting abundance and the contribution to the prediction is significantly more than would be expected by simply adding a random feature to the model.

\section{Discussion}
\label{sec:Discussion}

This work demonstrates that formal statistical inference procedures are possible within the context of supervised ensemble learning, even when individual base learners are difficult to analyze mathematically.  Demonstrating that ensembles built with subsamples can be viewed as U-statistics allows us to calculate limiting distributions for predictions and consistent estimation procedures for the parameters allow us to compute confidence intervals for predictions and formally test the significance of features.  Moreover, using the internal variance estimation procedure, we are able to do so at no additional computational cost.  In his controversial paper, \cite{breimantwocultures} contrasted traditional statistical data analysis models that emphasize interpretation with the modern algorithmic approach of machine learning where the primary goal is prediction and among the differences invoked was the statistician's concern for formalized inference.  Our hope is that this work be seen as something of a bridge between Breiman's two cultures.  

The distributions and procedures we discuss apply to a very general class of supervised learning algorithms.  We focus on bagging and random forests with tree-based base learners due to their popularity, but any supervised ensemble method that satisfies the conditions in Theorems \ref{subbaggingthm} and \ref{RFthm} will generate predictions that have these limiting distributions and the inference procedures can be carried out in the same way.  By the same reasoning, our procedures also make no restrictions on the tree-building method.

There are also some small modifications that can be made to our procedure which would still allow for an asymptotically normal limiting distribution.  First, we assumed that our subsamples were selected with replacement so that in theory, the same subsample could be selected multiple times.  This choice was based primarily on the fact that ensuring the exact subsample was not taken twice would typically require extra computational work.  However, even for relatively small datasets, the probability of selecting the same subsample more than once is very small, and not surprisingly, the limiting distributions remain identical when the subsamples are taken without replacement.  Furthermore, we also did not allow repeat observations within subsamples, which made the resulting estimator a U-statistic.  Had we selected the subsamples themselves with replacement, our estimator would be a V-statistic -- a closely related class of estimators introduced by \cite{vstatistics} -- and similar theory could be developed.  It is also worth pointing out that in practice, we always selected the same number of subsamples $m_n$ and subsample size $k_n$ for each prediction point $\predpoint$ but in theory, these could be chosen differently for each point of interest and the same can be said of the estimation parameters $\nxone$ and $\nmc$.  However, choosing different values of these parameters for different prediction points involves significantly more bookkeeping and we advise against it in practice.

Our approach also raises some issues.  Perhaps most obviously, the parameter in our inference procedures is the expected prediction generated by trees built in the prescribed fashion, as opposed to the true value of the underlying regression function, $F(\predpoint)$.  Though some tree-building methods have been shown to be consistent, the bias introduced may not be negligible so we have to be careful about interpreting our results.  It may be possible to employ a residual bootstrap to try and correct for bias and we plan to explore this in future work.  Another open question that we hope to address in future work is how to select the test points when testing feature significance.  In the eBird application, we randomly selected points from the training set, but it would be beneficial to investigate optimal strategies for selecting both the number and location of test points.  Finally, the distributional results we provide could potentially allow us to test more complex hypotheses about the structure of the underlying regression function, for example, the interaction between covariates. 





\vspace{4mm}

\section*{Acknowledgements}
This work was supported by the following grants:  NSF DEB-125619, NSF CDI Type II 1125098, and NSF DMS-1053252.  The authors would also like to thank the Lab of Ornithology at Cornell University for providing interesting data.



\newpage
\appendix
\section*{Appendix A.}
\label{AppendixA}

We present here the proofs of the theorems provided in Section~\ref{sec:TreesAsUStatistics}.  We begin with a lemma from \cite{Lee1990}.

\begin{lemma}
\label{LeeLemma}
(Lee 1990, Lemma A, page 201) Let $a_1, a_2, ...$ be a sequence of constants such that $\lim_{n \rightarrow \infty} \frac{1}{n} \sum_{i=1}^{n} a_i = 0$ and $\lim_{n \rightarrow \infty} \frac{1}{n} \sum_{i=1}^{n} a_{i}^{2} = \sigma^{2}$ and let the random variables $M_1, ..., M_n$ have a multinomial distribution, $multinomial(m_n; \frac{1}{m_n}, ..., \frac{1}{m_n})$.  Then as $m_n,n \rightarrow \infty$, the limiting distribution of 

\[
m_{n}^{-1/2} \sum_{i=1}^{n} a_i (M_i - {m_n}/n)
\]

\noindent is $\mathcal{N}(0,\sigma^2)$.
\end{lemma}

\vspace{5mm}

\noindent Additionally, it will be useful to have the limiting distribution of complete infinite order U-statistics, which we provide in the lemma below.

\vspace{5mm}

\begin{lemma}
\label{IOUSdistn}
Let $Z_1, Z_2, ... \iid F_Z$ and let $\uk$ be a complete, infinite order U-statistic with kernel $h_{k_n}$ satisfying Condition 1 and $\theta_{k_n} = \mathbb{E}h_{k_n}(Z_1, ..., Z_{k_n})$ such that $\mathbb{E}h_{k_n}^{2}(Z_1, ..., Z_{k_n}) \leq C < \infty$ for all $n$ and some constant $C$ and $\lim \frac{k_n}{\sqrt{n}} = 0$.  Then 

\[
\frac{\sqrt{n}(\uk - \theta_{k_n})}{\sqrt{k_{n}^{2} \zeta_{1,k_n}}} \indist \mathcal{N}(0,1).
\]

\end{lemma}

\vspace{5mm}

\noindent The proof of Lemma \ref{LeeLemma} is provided in \cite{Lee1990} page 201.  The proof of Lemma \ref{IOUSdistn} follows in exactly the same fashion as the proof of result \emph{(i)} in Theorem \ref{subbaggingthm} below.  We take advantage of these lemmas in the following proofs.

\vspace{10mm}
\noindent
{\bf Theorem 1} {\it Let $Z_1, Z_2, ... \iid F_Z$ and let $U_{n,k_n,m_n}$ be an incomplete, infinite order U-statistic with kernel $h_{k_n}$ that satisfies Condition 1.  Let $\theta_{k_n} = \mathbb{E}h_{k_n}(Z_1, ..., Z_{k_n})$ such that $\mathbb{E}h_{k_n}^{2}(Z_1, ..., Z_{k_n}) \leq C < \infty$ for all $n$ and some constant $C$, and let $\lim \frac{n}{m_n} = \alpha$.  Then as long as $\lim \frac{k_n}{\sqrt{n}} = 0$ and $\lim \zeta_{1,k_n} \neq 0$,
\begin{enumerate}[(i)]
\item if $\alpha = 0$, then $\frac{\sqrt{n}(U_{n,k_n,m_n} - \theta_{k_n})}{\sqrt{k_{n}^{2} \zeta_{1,k_n}}} \indist \mathcal{N}(0,1)$.
\item if $0 < \alpha < \infty$, then $\frac{\sqrt{m_n}(U_{n,k_n,m_n} - \theta_{k_n})}{\sqrt{\frac{k_{n}^{2}}{\alpha} \zeta_{1,k_n} + \zeta_{k_n,k_n}}} \indist \mathcal{N}(0,1)$.
\item if $\alpha = \infty$, then $\frac{\sqrt{m_n}(U_{n,k_n,m_n} - \theta_{k_n})}{\sqrt{\zeta_{k_n,k_n}}} \indist \mathcal{N}(0,1)$.
\end{enumerate}} 


\vspace{5mm}

\noindent
{\bf Proof:} \\

\noindent   \emph{(i)} \hspace{6mm} Suppose first that $\alpha = 0$.  In the interest of clarity, we follow the H\'ajek projection method discussed in \cite{vandervaart} chapters 11 and 12.  Define the H\'ajek projection of $\ukm - \theta_{k_n}$ as

\begin{align*}
\hatukm &= \sum_{i=1}^{n} \mathbb{E}(\ukm - \theta_{k_n} \; | \; Z_i) - (n-1) \mathbb{E}(\ukm - \theta_{k_n}) \\
&= \sum_{i=1}^{n} \mathbb{E}(\ukm - \theta_{k_n} \; | \; Z_i)
\end{align*}

\noindent so that for each term in the sum, we have

\begin{align}
\mathbb{E}(\ukm - \theta_{k_n} \; | \; Z_i) &= \mathbb{E} \left( \frac{1}{m_n} \sum_{\beta}h_{k_n}(Z_{\beta_1}, ..., Z_{\beta_{k_n}}) - \theta_{k_n} \; \bigg| \; Z_i \right) \notag \\ 
\label{OneSum} &=  \frac{1}{m_n} \sum_{\beta} \mathbb{E} \big(h_{k_n}(Z_{\beta_1}, ..., Z_{\beta_{k_n}}) - \theta_{k_n} \; \big| \; Z_i \big)
\end{align}

\noindent where, in keeping with the notation in \cite{vandervaart}, we let $\beta$ index the subsamples.  Define $h_{1,k_n}(x) = \mathbb{E}h_{k_n}(x, Z_2, ..., Z_{k_n}) - \theta_{k_n}$ and let $W$ be the number of subsamples that contain $i$.  Since we assume that the subsamples are selected uniformly at random with replacement, 

\[
W \sim Binom \left( m_n,\frac{\binom{n-1}{k_n-1}}{\binom{n}{k}} \right) 
\]

\noindent so we can rewrite (\ref{OneSum}) as 

\begin{align*}
&\frac{1}{m_n} \sum_{\beta} \mathbb{E} \bigg( \mathbb{E} (h_{k_n}(Z_{\beta_1}, ..., Z_{\beta_{k_n}}) - \theta_{k_n} | Z_i) \; \big| \; W \bigg) \\
&= \frac{1}{m_n} \mathbb{E} (W h_{1,k_n}(Z_i)) \\
&= \frac{1}{m_n} \left( m_n \left( \frac{\binom{n-1}{k_n-1}}{\binom{n}{k}} \right) h_{1,k_n}(Z_i) \right) \\
&= \frac{k_n}{n} h_{1,k_n}(Z_i)
\end{align*}

\noindent so that taking the sum yields

\[
\hatukm = \frac{k_n}{n} \sum_{i=1}^n h_{1,k_n}(Z_i).
\]

\noindent Now we establish the asymptotic normality of $\hatukm$.  Define the triangular array
\begin{align*}
& k_{n_1}h_{1,k_{n_1}}(Z_1),\; \dotsc \; , k_{n_1}h_{1,k_{n_1}}(Z_{n_1}) \\
& k_{n_1 + 1}h_{1,k_{n_1 + 1}}(Z_1),\; \; \; \dotsc \; \; \; \dotsc \; \; \;, k_{n_1 + 1}h_{1,k_{n_1 + 1}}(Z_{n_1 + 1}) \\
& \; \; \; .  \hspace{60mm}. \\
& \; \; \; .  \hspace{70mm}.  \\
& \; \; \; .   \hspace{80mm}.  \\
& k_{n_1 + j}h_{1,k_{n_1 + j}}(Z_1),\; \; \; \dotsc \; \; \; \dotsc \; \; \; \; \; \; \dotsc \; \; \; \dotsc \; \; \; \; \; \; \dotsc \; \; \; \dotsc \; \; \;, k_{n_1 + j}h_{1,k_{n_1 + j}}(Z_{n_1 + j}) \\
\end{align*}

\noindent so that for each variable in the array, we have

\[
\mathbb{E}(k_nh_{1,k_{n}}(Z_i)) = k_n(\theta_{k_n} - \theta_{k_n}) = 0
\]

\noindent and

\[
var(k_nh_{1,k_n}(Z_i)) = k_{n}^2 var(h_{1,k_n}(Z_i)) = k_{n}^2 \zeta_{1,k_n}
\]

\noindent and thus the row-wise sum of the variances is 

\[
s_{n}^2 = \sum_{i=1}^n var(k_n h_{1,k_n}(Z_i)) = nk_{n}^2 \zeta_{1,k_n}.
\]

%
%
%
%
\noindent For $\delta > 0$, the Lindeberg condition is given by 
\begin{align}
&\lim_{n \rightarrow \infty} \sum_{i=1}^n \frac{1}{n k_{n}^{2} \zeta_{1,k_n}} \int_{|k_n h_{1,k_n}(Z_i)| \geq \delta k_n \sqrt{n \zeta_{1,k_n}}} k_{n}^{2} h_{1,k_n}^{2}(Z_i) dP \nonumber \\
&= \lim_{n \rightarrow \infty} \sum_{i=1}^n \frac{1}{n \zeta_{1,k_n}} \int_{|h_{1,k_n}(Z_i)| \geq \delta \sqrt{n \zeta_{1,k_n}}} h_{1,k_n}^{2}(Z_i) dP \nonumber \\
&\leq \lim_{n \rightarrow \infty} \max_{1 \leq i \leq n} \frac{1}{\zeta_{1,k_n}} \int_{|h_{1,k_n}(Z_i)| \geq \delta \sqrt{n \zeta_{1,k_n}}} h_{1,k_n}^{2}(Z_i) dP \nonumber \\
&\label{LCT1} = \lim_{n \rightarrow \infty} \frac{1}{\zeta_{1,k_n}} \int_{|h_{1,k_n}(Z_1)| \geq \delta \sqrt{n \zeta_{1,k_n}}} h_{1,k_n}^{2}(Z_1) dP \\ \nonumber
&= 0
\end{align}

\noindent by Condition 1, and thus the Lindeberg condition is satisfied.  Thus, by the Lindeberg-Feller central limit theorem, 

\[
\frac{\sum_j k_{n}h_{1,k_{n}}(Z_j)}{s_n} \indist \mathcal{N}(0,1) 
\]

\noindent or, rewriting,

\[
\frac{\sqrt{n} \hatukm}{\sqrt{k_{n}^2 \zeta_{1,k_n}}} \indist \mathcal{N}(0,1).
\]

\noindent Now, we need to compare the limiting variance ratio of $\ukm$ and its H\'ajek projection $\hatukm$.  For incomplete U-statistics of fixed rank, \cite{blom} showed that the variance of the incomplete U-statistic $U_m$ consisting of $m$ subsamples selected uniformly at random with replacement is given by 

\[
\frac{\zeta_k}{m_n} + \left( 1 - \frac{1}{m_n} \right) var(U)
\]

\noindent where $U$ is the complete U-statistic analogue.  Extending this result to our situation where $k$ and $m$ may both depend on $n$, we have

\[
var(\ukm) = \frac{\zeta_{k_n,k_n}}{m_n} + \left( 1 - \frac{1}{m_n} \right) var(U_{n,k_n})
\]

\noindent where the variance of the complete U-statistic $U_{n,k_n}$ is given by

\[
\sum_{c=1}^{k_n} \frac{k_{n}!^2}{c!(k_n-c)!^2} \frac{(n-k_n)(n-k_n-1) \dotsb (n-2k_n+c+1)}{n(n-1) \dotsb (n-k_n+1)} \zeta_{c,k_n}.
\]

\noindent The details of this calculation are described in \cite{vandervaart} page 163.  Thus, looking at the limit of the variance ratio, we have 

\begin{align*}
\lim_{n \rightarrow \infty} \left( \frac{var(\ukm)}{var(\hatukm)} \right) &= \lim_{n \rightarrow \infty} \left( \frac{\frac{\zeta_{k_n,k_n}}{m_n} + \left( 1 - \frac{1}{m_n} \right) var(U_{n,k_n})}{\frac{k_n^2}{n} \zeta_{1,k_n}} \right) \notag \\
&= \lim_{n \rightarrow \infty} \left( \frac{n \; \zeta_{k_n,k_n}}{m_n \; k_{n}^{2} \; \zeta_{1,k_n}} \right) + \lim_{n \rightarrow \infty} \left( 1 - \frac{1}{m_n} \right) \lim_{n \rightarrow \infty} \left( \frac{n \; var(U_{n,k_n})}{k_{n}^{2} \; \zeta_{1,k_n}} \right) \notag \\
&= 0 + \lim_{n \rightarrow \infty} \left( \frac{n \; var(U_{n,k_n})}{k_{n}^{2} \; \zeta_{1,k_n}} \right) \\
&= \lim_{n \rightarrow \infty} \left( \frac{\sum_{c=1}^{k_n} \frac{k_{n}!^2}{c!(k_n-c)!^2} \frac{(n-k_n)(n-k_n-1) \dotsb (n-2k_n+c+1)}{(n-1) \dotsb (n-k_n+1)} \zeta_{c,k_n}}{k_n^2 \zeta_{1,k_n}} \right) \notag \\
&= \lim_{n \rightarrow \infty} \left( \frac{\frac{k_{n}!^2}{(k_n-1)!^2} \frac{(n-k_n)(n-k_n-1) \dotsb (n-2k_n+2)}{(n-1) \dotsb (n-k_n+1)} \zeta_{1,k_n}}{k_n^2 \zeta_{1,k_n}} \right) \\
&= \lim_{n \rightarrow \infty} \left( \frac{(n-k_n)(n-k_n-1) \dotsb (n-2k_n+2)}{(n-1) \dotsb (n-k_n+1)} \right) \notag \\
&= 1. \notag
\end{align*}

\noindent Note that in the second line, $\lim \left( \frac{n \; \zeta_{k_n,k_n}}{m_n \; k_{n}^{2} \; \zeta_{1,k_n}} \right) = 0$ since $\frac{n}{m_n} \rightarrow 0$, $\zeta_{1,k_n} \nrightarrow 0$ by assumption, and $\zeta_{k_n,k_n} \nrightarrow \infty$ since $\mathbb{E}h_{k_n}^{2}(Z_1, ..., Z_{k_n})$ is bounded.  Finally, by Slutsky's Theorem and Theorem 11.2 in \cite{vandervaart}, we have

\begin{align*}
\frac{\sqrt{n}(\ukm - \theta_{k_n})}{\sqrt{k^{2}_{n}\zeta_{1,k_n}}} &= \frac{\sqrt{n}(\ukm - \theta_{k_n} - \hatukm + \hatukm)}{\sqrt{k^{2}_{n}\zeta_{1,k_n}}} \\
&= \frac{\sqrt{n}(\ukm - \theta_{k_n} - \hatukm)}{\sqrt{k^{2}_{n}\zeta_{1,k_n}}} + \frac{\sqrt{n} \hatukm}{\sqrt{k^{2}_{n}\zeta_{1,k_n}}}
\end{align*}

\noindent where $\sqrt{n}(\ukm - \theta_{k_n} - \hatukm) / {\sqrt{k^{2}_{n}\zeta_{1,k_n}}} \inprob 0$ and 

\[
\frac{\sqrt{n} \hatukm}{\sqrt{k^{2}_{n}\zeta_{1,k_n}}} \indist \mathcal{N}(0,1)
\]

\noindent so that 

\[
\frac{\sqrt{n}(\ukm - \theta_{k_n})}{\sqrt{k^{2}_{n}\zeta_{1,k_n}}} \indist \mathcal{N}(0,1)
\]

\noindent as desired. $\square$

\vspace{15mm}

\noindent \emph{(ii)} $\&$ \emph{(iii)} \hspace{6mm} Now suppose $\alpha > 0$.  We follow the proof technique in \cite{Lee1990} page 200 which is based on the work of \cite{Janson1984}.  Let $\mathcal{S}_{(n,k_n)} = \{ S_i:  i = 1, ..., \binom{n}{k_n}\}$ denote the set of all possible subsamples of size $k_n$.  In the following work, we use the notation $(n,k_n)$ in place of $\binom{n}{k_n}$ in subscripts and summation notation.  Consider the random vector $M_{n,k_n}= (M_{S_1}, ..., M_{S_{(n,k_n)}})$, where the $i^{th}$ element denotes the number of times the $i^{th}$ subsample appears in $\ukm$.  Since the subsamples are selected uniformly at random with replacement, $M_{n,k_n} \sim multinomial(m_n; \frac{1}{\binom{n}{k_n}}, ..., \frac{1}{\binom{n}{k_n}})$.  Let $\phi_{n,k_n,m_n}$ be the characteristic function of $\sqrt{m_n}(\ukm - \theta_{k_n})$ and let $\phi$ denote the limiting characteristic function of $\sqrt{n}(\uk - \theta_{k_n})$ where $\uk$ is the corresponding complete U-statistic.  Additionally, let $\phi^{(M)}_{n,k_n,m_n}$ be the characteristic function of the random variable

\[
m_{n}^{-1/2} \sum_{i=1}^{(n,k)} \Big( M_{S_i} - \frac{m_n}{\binom{n}{k_n}} \Big) \Big( h_{k_n}(S_i) - \theta_{k_n} \Big) \Big| Z_1, ..., Z_n.
\]

\noindent Then we have

\begin{align*}
\phi_{n,k_n,m_n}(t) &= \mathbb{E} \left( exp \left[ it \sqrt{m_n}(\ukm - \theta_{k_n}) \right] \right) \\
&= \mathbb{E} \left( exp \left[ it m_{n}^{-1/2} \left( \sum_{i=1}^{(n,k_n)} M_{S_i}(h_{k_n}(S_i) - \theta_{k_n}) \right) \right] \right) \\
&= \mathbb{E} \left( \mathbb{E} \left( exp \left[ it m_{n}^{-1/2} \left( \sum_{i=1}^{(n,k_n)} M_{S_i}(h_{k_n}(S_i) - \theta_{k_n}) \right) \right] \middle \vert Z_1, ..., Z_n \right) \right)\\
&= \mathbb{E} \Bigg( \mathbb{E} \Bigg( exp \Bigg[ it m_{n}^{-1/2} \Bigg( \sum_{i=1}^{(n,k_n)} \bigg( M_{S_i} + \frac{m_n}{\binom{n}{k_n}} -  \frac{m_n}{\binom{n}{k_n}} \bigg) \\
&\hspace{53mm} \times \big(h_{k_n}(S_i) - \theta_{k_n}\big) \Bigg) \Bigg] \; \Bigg| \; Z_1, ..., Z_n \Bigg) \Bigg)\\
&= \mathbb{E} \Bigg( \mathbb{E} \Bigg( exp \Bigg[ it m_{n}^{-1/2} \left( \sum_{i=1}^{(n,k_n)} \frac{m_n}{\binom{n}{k_n}} (h_{k_n}(S_i) - \theta_{k_n}) \right) \Bigg] \\
&\hspace{8mm} \times exp \Bigg[ it m_{n}^{-1/2} \left( \sum_{i=1}^{(n,k)} \left( M_{S_i} -  \frac{m_n}{\binom{n}{k_n}} \right)(h_{k_n}(S_i) - \theta_{k_n}) \right) \Bigg] \bigg| Z_1, ..., Z_n \Bigg) \Bigg)\\
&= \mathbb{E} \Bigg( exp \Bigg[ it m_{n}^{-1/2} \left( \sum_{i=1}^{(n,k_n)} \frac{m_n}{\binom{n}{k_n}} (h_{k_n}(S_i) - \theta_{k_n}) \right) \Bigg] \\
&\hspace{8mm} \times \mathbb{E} \Bigg( exp \Bigg[ it m_{n}^{-1/2} \left( \sum_{i=1}^{(n,k_n)} \left( M_{S_i} -  \frac{m_n}{\binom{n}{k_n}} \right)(h_{k_n}(S_i) - \theta_{k_n}) \right) \Bigg] \bigg| Z_1, ..., Z_n \Bigg) \Bigg)\\
&= \mathbb{E} \Bigg( exp \bigg[ it \sqrt{m_n} \uk \bigg] \hspace{2mm} \phi_{n,k_n,m_n}^{(M)}(t) \Bigg) \\
\end{align*}

\noindent and now, taking limits, we have

\begin{align*}
\lim_{n \rightarrow \infty} \phi_{n,k_n,m_n}(t) &= \lim_{n \rightarrow \infty} \mathbb{E} \Bigg( exp \bigg[ it \sqrt{m_n} \uk \bigg] \hspace{2mm} \phi_{n,k_n,m_n}^{(M)}(t) \Bigg) \\
&= \mathbb{E} \Bigg( \lim_{n \rightarrow \infty} exp \bigg[ it \sqrt{m_n} \uk \bigg] \hspace{2mm} \lim_{n \rightarrow \infty} \phi_{n,k_n,m_n}^{(M)}(t) \Bigg) \\
\end{align*}

\noindent so that by the preceeding lemmas,

\begin{align*}
\lim_{n \rightarrow \infty} \phi_{n,k_n,m_n}(t) &= \lim_{n \rightarrow \infty} \mathbb{E} \Bigg( exp \bigg[ it \sqrt{m_n} \uk \bigg] \Bigg) exp \Bigg[ - t^2 \zeta_{k_n,k_n} / 2 \Bigg] \\
&= \lim_{n \rightarrow \infty} \mathbb{E} \Bigg( exp \bigg[ it \Big( \frac{\sqrt{m_n}}{\sqrt{n}} \Big) \sqrt{n} \uk \bigg] \Bigg) exp \Bigg[ - t^2 \zeta_{k_n,k_n} / 2 \Bigg] \\
&=\phi(\alpha^{-1/2} t) exp \Bigg[ - t^2 \zeta_{k_n,k_n} / 2 \Bigg] \\
&= exp \Bigg[ - (t \alpha^{-1/2})^2 k_{n}^{2} \zeta_{1,k_n} / 2 \Bigg] exp \Bigg[ - t^2 \zeta_{k_n,k_n} / 2 \Bigg] \\
&= exp \Bigg[ -t^2 \Big( \frac{k_{n}^{2}}{\alpha} \zeta_{1,k_n}  + \zeta_{k_n,k_n} \Big) / 2 \Bigg] \\
\end{align*}

\noindent which is the characteristic function of a Normal distribution with mean 0 and variance $\frac{k_{n}^{2}}{\alpha} \zeta_{1,k_{n}}  + \zeta_{k_n,k_n}$.  Note that when $\alpha = \infty$, the first term in the variance is 0, so the limiting variance reduces to $\zeta_{k_n,k_n}$, as desired. $\blacksquare$

\vspace{10mm}
\noindent \textbf{Proposition 1:  }  \emph{For a bounded regression function $F$, if there exists a constant $c$ such that for all $k_n \geq 1$, }
\begin{align*}
\big| h((\bm{X}_{1},Y_{1}), ..., (\bm{X}_{k_n},Y_{k_n}), (\bm{X}_{k_n+1},Y_{k_n+1})) - h((\bm{X}_{1},Y_{1}), ..., (\bm{X}_{k_n},&Y_{k_n}), (\bm{X}_{k_n+1},Y^{*}_{k_n+1})) \big| \\ 
\leq c \big| Y_{k_n+1} - Y^{*}_{k_n+1} \big|
\end{align*}
\noindent \emph{where $Y_{k_n+1} = F(\bm{X}_{k_n+1}) + \epsilon_{k_n+1}$, $Y^{*}_{k_n+1} = F(\bm{X}_{k_n+1}) + \epsilon^{*}_{k_n+1}$, and where $\epsilon_{k_n+1}$ and $\epsilon^{*}_{k_n+1}$ are i.i.d. with exponential tails, then Condition 1 is satisfied.}

\vspace{5mm}

\noindent
{\bf Proof}. First consider the particular $Y^{*}_{j} = F(\bm{X}_{j})$.  Since $F$ is bounded, we can define 

\[
\sup_{\bm{X}_j \in \mathcal{X}} \big| F(\bm{X}_j) \big| \leq M < \infty
\]

\noindent and since tree-based predictions cannot fall outside the range of responses in the training set, $\big| h\big((\bm{X}_{1},Y_{1}^{*}), ...,(\bm{X}_{k_n},Y_{k_n}^{*})\big) \big| \leq M$ for all possible $\bm{X}_1, ..., \bm{X}_{k_n}$.  Furthermore, by the Lipschitz condition,

\[
\big| h\big((\bm{X}_{1},Y_{1}^{*}), ...,(\bm{X}_{k_n},Y_{k_n}^{*})\big) - h\big((\bm{X}_{1},Y_{1}^{*}), (\bm{X}_{2},Y_{2}), ...,(\bm{X}_{k_n},Y_{k_n})\big) \big| \leq c \sum_{j=2}^{k_n} \big| \epsilon_j \big|
\]

\noindent and thus, applying Jensen's Inequality, we have

\begin{align*}
&\sup_{\bm{X}_1 \in \mathcal{X}} \big| h_{1,k_n}\big( (\bm{X}_1, Y_{1}^{*}) \big) \big| \\
&= \sup_{\bm{X}_1 \in \mathcal{X}} \Bigg| \int h\big((\bm{X}_{1},Y_{1}^{*}), (\bm{X}_{2},Y_{2}), ...,(\bm{X}_{k_n},Y_{k_n})\big) - h\big((\bm{X}_{1},Y_{1}^{*}), ...,(\bm{X}_{k_n},Y_{k_n}^{*})\big) \\
&\hspace{70mm} + h\big((\bm{X}_{1},Y_{1}^{*}), ...,(\bm{X}_{k_n},Y_{k_n}^{*})\big) dP - \theta_{k_n} \Bigg| \\
&\leq \sup_{\bm{X}_1 \in \mathcal{X}} \int \Big| h\big((\bm{X}_{1},Y_{1}^{*}), (\bm{X}_{2},Y_{2}), ...,(\bm{X}_{k_n},Y_{k_n})\big) - h\big((\bm{X}_{1},Y_{1}^{*}), ...,(\bm{X}_{k_n},Y_{k_n}^{*})\big) \Big| dP \\
&\hspace{70mm} + \sup_{\bm{X}_1 \in \mathcal{X}} h\big((\bm{X}_{1},Y_{1}^{*}), ...,(\bm{X}_{k_n},Y_{k_n}^{*})\big) - \theta_{k_n} \\
&\leq ck_n \mathbb{E} \big| \epsilon_1 \big| + M - \theta_{k_n}. \\
\end{align*}

\noindent Now, define the set of interest in the Lindeberg condition as $A_n$ so that we may write

\begin{align*}
A_n &= \bigg\{ \Big| h_{1,k_n}\big( (\bm{X}_1, Y_{1}) \big) \Big| \geq \delta \sqrt{n \zeta_{1,k_n}} \bigg\} \\
&= \bigg\{ \Big| h_{1,k_n}\big( (\bm{X}_1, Y_{1}) \big) - h_{1,k_n}\big( (\bm{X}_1, Y_{1}^{*}) \big) + h_{1,k_n}\big( (\bm{X}_1, Y_{1}^{*}) \big) \Big| \geq \delta \sqrt{n \zeta_{1,k_n}} \bigg\} \\
&\subseteq \bigg\{ \Big| h_{1,k_n}\big( (\bm{X}_1, Y_{1}) \big) - h_{1,k_n}\big( (\bm{X}_1, Y_{1}^{*}) \big) \Big| \geq \delta \sqrt{n \zeta_{1,k_n}}  + \Big| h_{1,k_n}\big( (\bm{X}_1, Y_{1}^{*}) \big) \Big| \bigg\} \\
&\subseteq \bigg\{ \big| \epsilon_1 \big| \geq \frac{1}{c}\left(\delta \sqrt{n \zeta_{1,k_n}} + M - \theta_{k_n} \right) + k_n \mathbb{E} \big| \epsilon_1 \big| \bigg\} \\
&=: A_{n}^{*}
\end{align*}

\noindent Finally, continuing from equation (\ref{LCT1}) of Theorem 1,

\begin{align*}
&\lim_{n \rightarrow \infty} \frac{1}{\zeta_{1,k_n}} \int_{A_n} h_{1,k_n}^{2} \big( (\bm{X}_1,Y_1) \big) dP \\
&= \lim_{n \rightarrow \infty} \frac{1}{\zeta_{1,k_n}} \int_{A_n} \left( h_{1,k_n} \big( (\bm{X}_1,Y_1) \big) - h_{1,k_n} \big( (\bm{X}_1,Y_{1}^{*}) + h_{1,k_n} \big( (\bm{X}_1,Y_{1}^{*}) \right)^2 dP \\ 
&\leq \lim_{n \rightarrow \infty} \frac{2}{\zeta_{1,k_n}} \int_{A_n} \left( h_{1,k_n} \big( (\bm{X}_1,Y_1) \big) - h_{1,k_n} \big( (\bm{X}_1,Y_{1}^{*}) \right)^2 dP \\
&\hspace{50mm} + \lim_{n \rightarrow \infty} \frac{2}{\zeta_{1,k_n}} \int_{A_n} h_{1,k_n}^{2} \big( (\bm{X}_1,Y_{1}^{*}) dP \\
&\leq \lim_{n \rightarrow \infty} \frac{2}{\zeta_{1,k_n}} \int_{A_{n}^{*}} c^2 \epsilon_{1}^{2} dP  + \lim_{n \rightarrow \infty} \frac{2}{\zeta_{1,k_n}} P(A_{n}^{*}) \left( ck_n \mathbb{E} \big| \epsilon_1 \big| + M -\theta_{k_n} \right)^2 \\
&= \lim_{n \rightarrow \infty} \frac{2}{\zeta_{1,k_n}} P \left[ \big| \epsilon_1 \big| \geq \frac{1}{c} \left( \sqrt{n \zeta_{1,k_n}} + M -\theta_{k_n} \right) + k \mathbb{E} \big| \epsilon_1 \big| \right] \\
&\hspace{50mm} \times \left( ck_n \mathbb{E} \big| \epsilon_1 \big| + M - \theta_{k_n} \right)^2 \\
&= 0 \\
\end{align*}

\noindent as desired, so long as $n P \left( \big| \epsilon_1 \big| > \sqrt{n} \right) \rightarrow 0$ -- which is the case with exponential tails -- and $k_n = o(\sqrt{n})$. $\blacksquare$

\vspace{10mm}
\noindent
{\bf Theorem 2} {\it Let $U_{\omega; n,k_n,m_n}$ be a random kernel U-statistic of the form defined in equation (\ref{RFestimator}) such that $U_{\omega; n,k_n,m_n}^{*}$ satisfies Condition 1 and suppose that $\mathbb{E}h_{k_n}^{2}(Z_1, ..., Z_{k_n}) < \infty$ for all $n$, $\lim \frac{k_n}{\sqrt{n}} = 0$, and $\lim \frac{n}{m_n} = \alpha$.  Then, letting $\beta$ index the subsamples, so long as $\lim \zeta_{1,k_n} \neq 0$ and 

\[
\lim_{n \rightarrow \infty} \mathbb{E} \left( h_{k_n}^{(\omega)}(Z_{\beta_{1}}, ..., Z_{\beta_{k_n}}) - \mathbb{E}_{\omega} h_{k_n}^{(\omega)}(Z_{\beta_{1}}, ..., Z_{\beta_{k_n}}) \right)^2 \neq \infty,
\]

\noindent $U_{\omega; n,k_n,m_n}$ is asymptotically normal and the limiting distributions are the same as those provided in Theorem~\ref{subbaggingthm}.} 

\vspace{5mm}

\noindent
{\bf Proof}. We begin with the case where $\alpha = 0$ and we make use of this result in the proof of the case where $\alpha>0$.  As in Section \ref{subsec:rf}, define $U_{\omega; n,k_n,m_n}^{*} = \mathbb{E}_{\omega} U_{\omega; n,k_n,m_n}$.  We have

\begin{align*}
\mathbb{E} & (U_{\omega; n,k_n,m_n} - U_{\omega; n,k_n,m_n}^{*})^2 \notag \\
&= \mathbb{E} \left[ \left( \frac{1}{m_n} \sum_{\beta} h_{k_n}^{(\omega)}(Z_{\beta_{1}}, ..., Z_{\beta_{k_n}}) - \mathbb{E}_{\omega} \left( \frac{1}{m_n} \sum_{\beta} h_{k_n}^{(\omega)}(Z_{\beta_{1}}, ..., Z_{\beta_{k_n}}) \right) \right)^2 \right] \notag \\
&= \mathbb{E} \left[ \frac{1}{m_n^2}\left( \sum_{\beta} h_{k_n}^{(\omega)}(Z_{\beta_{1}}, ..., Z_{\beta_{k_n}}) - \mathbb{E}_{\omega} \left( \sum_{\beta} h_{k_n}^{(\omega)}(Z_{\beta_{1}}, ..., Z_{\beta_{k_n}}) \right) \right)^2 \right] \notag \\
&= \mathbb{E} \frac{1}{m_n^2}\left( \sum_{\beta} h_{k_n}^{(\omega)}(Z_{\beta_{1}}, ..., Z_{\beta_{k_n}}) - \left( \sum_{\beta} \mathbb{E}_{\omega} h_{k_n}^{(\omega)}(Z_{\beta_{1}}, ..., Z_{\beta_{k_n}}) \right) \right)^2 \notag \\
&= \mathbb{E} \left[ \frac{1}{m_n^2} \sum_{\beta} \left( h_{k_n}^{(\omega)}(Z_{\beta_{1}}, ..., Z_{\beta_{k_n}}) - \mathbb{E}_{\omega} h_{k_n}^{(\omega)}(Z_{\beta_{1}}, ..., Z_{\beta_{k_n}}) \right)^2 \right] \notag \\
&\hspace{3mm} + \mathbb{E} \Bigg[ \frac{1}{m_n^2} \sum_{\beta_{i} \neq \beta_{j}} \left( h_{k_n}^{(\omega)}(Z_{\beta_{i_{1}}}, ..., Z_{\beta_{i_{k_n}}}) - \mathbb{E}_{\omega} h_{k_n}^{(\omega)}(Z_{\beta_{i_{1}}}, ..., Z_{\beta_{i_{k_n}}}) \right) \notag \\
&\hspace{50mm} \times \; \left( h_{k_n}^{(\omega)}(Z_{\beta_{j_{1}}}, ..., Z_{\beta_{j_{k_n}}}) - \mathbb{E}_{\omega} h_{k_n}^{(\omega)}(Z_{\beta_{j_{1}}}, ..., Z_{\beta_{j_{k_n}}}) \right) \Bigg] \notag \\
\end{align*}

\noindent We focus now on the second term, involving the cross terms with different subsamples and randomization parameters.  Splitting apart the expectation and moving the expectation with respect to $\omega$ inside, we can write the second term as

\begin{align*}
&\mathbb{E} \Bigg[ \frac{1}{m_n^2} \sum_{\beta_{i} \neq \beta_{j}} \left( h_{k_n}^{(\omega)}(Z_{\beta_{i_{1}}}, ..., Z_{\beta_{i_{k_n}}}) - \mathbb{E}_{\omega} h_{k_n}^{(\omega)}(Z_{\beta_{i_{1}}}, ..., Z_{\beta_{i_{k_n}}}) \right) \\
&\hspace{35mm} \times \left( h_{k_n}^{(\omega)}(Z_{\beta_{j_{1}}}, ..., Z_{\beta_{j_{k_n}}}) - \mathbb{E}_{\omega} h_{k_n}^{(\omega)}(Z_{\beta_{j_{1}}}, ..., Z_{\beta_{j_{k_n}}}) \right) \Bigg] \notag \\
&= \mathbb{E}_{\bm{X}} \Bigg[ \frac{1}{m_n^2} \sum_{\beta_{i} \neq \beta_{j}} \mathbb{E}_{\omega} \left( h_{k_n}^{(\omega)}(Z_{\beta_{i_{1}}}, ..., Z_{\beta_{i_{k_n}}}) - \mathbb{E}_{\omega} h_{k_n}^{(\omega)}(Z_{\beta_{i_{1}}}, ..., Z_{\beta_{i_{k_n}}}) \right) \\
&\hspace{15mm} \times \mathbb{E}_{\omega} \left( h_{k_n}^{(\omega)}(Z_{\beta_{j_{1}}}, ..., Z_{\beta_{j_{k_n}}}) - \mathbb{E}_{\omega} h_{k_n}^{(\omega)}(Z_{\beta_{j_{1}}}, ..., Z_{\beta_{j_{k_n}}}) \right) \Bigg] \notag \\ 
&= \mathbb{E}_{\bm{X}} \Bigg[ \frac{1}{m_n^2} \sum_{\beta_{i} \neq \beta_{j}} \left( \mathbb{E}_{\omega} h_{k_n}^{(\omega)}(Z_{\beta_{i_{1}}}, ..., Z_{\beta_{i_{k_n}}}) - \mathbb{E}_{\omega} h_{k_n}^{(\omega)}(Z_{\beta_{i_{1}}}, ..., Z_{\beta_{i_{k_n}}}) \right) \notag \\
&\hspace{15mm} \times \left( \mathbb{E}_{\omega} h_{k_n}^{(\omega)}(Z_{\beta_{j_{1}}}, ..., Z_{\beta_{j_{k_n}}}) - \mathbb{E}_{\omega} h_{k_n}^{(\omega)}(Z_{\beta_{j_{1}}}, ..., Z_{\beta_{j_{k_n}}}) \right) \Bigg] \notag \\ 
&= \mathbb{E}_{\bm{X}} \Bigg[ \frac{1}{m_n^2} \sum_{\beta_{i} \neq \beta_{j}} 0 \times 0 \Bigg] \notag \\
&= 0 \notag 
\end{align*}

\noindent and thus we need only investigate the first term.  We have 

\begin{align*}
\mathbb{E} & \left[ \frac{1}{m_n^2} \sum_{\beta} \left( h_{k_n}^{(\omega)}(Z_{\beta_{1}}, ..., Z_{\beta_{k_n}}) - \mathbb{E}_{\omega} h_{k_n}^{(\omega)}(Z_{\beta_{1}}, ..., Z_{\beta_{k_n}}) \right)^2 \right] \\
&= \frac{1}{m_n^2} \sum_{\beta} \mathbb{E} \left( h_{k_n}^{(\omega)}(Z_{\beta_{1}}, ..., Z_{\beta_{k_n}}) - \mathbb{E}_{\omega} h_{k_n}^{(\omega)}(Z_{\beta_{1}}, ..., Z_{\beta_{k_n}}) \right)^2 \\
&= \frac{1}{m_n^2} m_n \mathbb{E} \left( h_{k_n}^{(\omega)}(Z_{\beta_{1}}, ..., Z_{\beta_{k_n}}) - \mathbb{E}_{\omega} h_{k_n}^{(\omega)}(Z_{\beta_{1}}, ..., Z_{\beta_{k_n}}) \right)^2 \\
&= \frac{1}{m_n} \mathbb{E} \left( h_{k_n}^{(\omega)}(Z_{\beta_{1}}, ..., Z_{\beta_{k_n}}) - \mathbb{E}_{\omega} h_{k_n}^{(\omega)}(Z_{\beta_{1}}, ..., Z_{\beta_{k_n}}) \right)^2. \\
\end{align*} 

\noindent Putting all of this together, we have 

\begin{align*}
&\lim_{n \rightarrow \infty} \mathbb{E} \left( \frac{\sqrt{n}\left(U_{\omega; n,k_n,m_n} - U_{\omega; n,k_n,m_n}^{*}\right)}{\sqrt{k_{n}^{2} \zeta_{1,k_n}}} \right)^2 \\  
\label{rfFinalLim} &= \lim_{n \rightarrow \infty} \frac{n}{m_n} \frac{1}{k_{n}^{2} \zeta_{1,k_n}} \mathbb{E} \left( h_{k_n}^{(\omega)}(Z_{\beta_{1}}, ..., Z_{\beta_{k_n}}) - \mathbb{E}_{\omega} h_{k_n}^{(\omega)}(Z_{\beta_{1}}, ..., Z_{\beta_{k_n}}) \right)^2 \\
&=0 
\end{align*}

\noindent so long as $k_{n}^{2} \zeta_{1,k_n} \nrightarrow 0$ and $\lim_{n \rightarrow \infty} \mathbb{E} \left( h_{k_n}^{(\omega)}(Z_{\beta_{1}}, ..., Z_{\beta_{k_n}}) - \mathbb{E}_{\omega} h_{k_n}^{(\omega)}(Z_{\beta_{1}}, ..., Z_{\beta_{k_n}}) \right)^2 \neq \infty$, as desired.   $\square$

\vspace{15mm}

Now we handle the case where $\alpha>0$.  First note that when $k_n = 1$ for all $n$, this reduces to simply averaging over an $i.i.d.$ sample and thus asymptotic normality can be obtained via the classic central limit theorem so assume that eventually $k_n>1$.  The remaining steps in this proof are nearly identical to the proof of results \emph{(ii)} and \emph{(iii)} of Theorem \ref{subbaggingthm}.  Again, let $\mathcal{S}_{(n,k_n)} = \{ S_i:  i = 1, ..., \binom{n}{k_n}\}$ denote the set of all possible subsamples of size $k_n$ and let $M_{n,k_n}= (M_{S_1}, ..., M_{S_{(n,k_n)}})$ denote the random vector that counts the number of times each subsample appears so that $M_{n,k_n} \sim multinomial(m_n; \frac{1}{\binom{n}{k_n}}, ..., \frac{1}{\binom{n}{k_n}})$ since we assume the subsamples are selected uniformly at random with replacement.  Define $\theta_{k_n}^{*} = \mathbb{E} U_{\omega; n,k_n,m_n}^{*}$, let $\phi_{n,k_n,m_n}$ be the characteristic function of $\sqrt{m_n}(U_{\omega; n,k_n,m_n} - \theta_{k_n}^{*})$, and let $\phi$ denote the limiting characteristic function of $\sqrt{n}(U_{\omega; n,k_n,m_n}^{*} - \theta_{k_n}^{*})$.  Finally, let $\phi^{(M)}_{n,k_n,m_n}$ be the characteristic function of the random variable

\[
m_{n}^{-1/2} \sum_{i=1}^{(n,k)} \Big( M_{S_i} - \frac{m_n}{\binom{n}{k_n}} \Big) \Big( \randkern(S_i) - \theta_{k_n}^{*} \Big) \Big| Z_1, ..., Z_n, \omega.
\]

\noindent Then we have

\begin{align*}
\phi_{n,k_n,m_n}(t) &= \mathbb{E} \left( exp \left[ it \sqrt{m_n}(U_{\omega; n,k_n,m_n} - \theta_{k_n}^{*}) \right] \right) \\
&= \mathbb{E} \left( exp \left[ it m_{n}^{-1/2} \left( \sum_{i=1}^{(n,k_n)} M_{S_i}(\randkern(S_i) - \theta_{k_n}^{*}) \right) \right] \right) \\
&= \mathbb{E} \left( \mathbb{E} \left( exp \left[ it m_{n}^{-1/2} \left( \sum_{i=1}^{(n,k_n)} M_{S_i}(\randkern(S_i) - \theta_{k_n}^{*}) \right) \right] \middle \vert Z_1, ..., Z_n,\omega \right) \right)\\
&= \mathbb{E} \Bigg( \mathbb{E} \Bigg( exp \Bigg[ it m_{n}^{-1/2} \Bigg( \sum_{i=1}^{(n,k_n)} \left( M_{S_i} + \frac{m_n}{\binom{n}{k_n}} -  \frac{m_n}{\binom{n}{k_n}} \right) \\
&\hspace{67mm} \times \big(\randkern(S_i) - \theta_{k_n}^{*}\big) \Bigg) \Bigg] \; \Bigg| \; Z_1, ..., Z_n,\omega \Bigg) \Bigg)\\
&= \mathbb{E} \Bigg( \mathbb{E} \Bigg( exp \Bigg[ it m_{n}^{-1/2} \left( \sum_{i=1}^{(n,k_n)} \frac{m_n}{\binom{n}{k_n}} (\randkern(S_i) - \theta_{k_n}^{*}) \right) \Bigg] \\
&\hspace{10.5mm} \times exp \Bigg[ it m_{n}^{-1/2} \left( \sum_{i=1}^{(n,k)} \left( M_{S_i} -  \frac{m_n}{\binom{n}{k_n}} \right)(\randkern(S_i) - \theta_{k_n}^{*}) \right) \Bigg] \bigg| Z_1, ..., Z_n,\omega \Bigg) \Bigg)\\
&= \mathbb{E} \Bigg( exp \Bigg[ it m_{n}^{-1/2} \left( \sum_{i=1}^{(n,k_n)} \frac{m_n}{\binom{n}{k_n}} (\randkern(S_i) - \theta_{k_n}^{*}) \right) \Bigg] \\
&\hspace{3mm} \times \mathbb{E} \Bigg( exp \Bigg[ it m_{n}^{-1/2} \left( \sum_{i=1}^{(n,k_n)} \left( M_{S_i} -  \frac{m_n}{\binom{n}{k_n}} \right)(\randkern(S_i) - \theta_{k_n}^{*}) \right) \Bigg] \bigg| Z_1, ..., Z_n,\omega \Bigg) \Bigg)\\
&= \mathbb{E} \Bigg( exp \Bigg[ it m_{n}^{-1/2} \Bigg( \sum_{i=1}^{(n,k_n)} \frac{m_n}{\binom{n}{k_n}} \bigg( \randkern(S_i) - \mathbb{E}_{\omega} \randkern(S_i) \\
&\hspace{69mm} + \mathbb{E}_{\omega} \randkern(S_i) - \theta_{k_n}^{*} \bigg) \Bigg) \Bigg] \hspace{2mm} \phi_{n,k_n,m_n}^{(M)}(t) \Bigg) \\
&= \mathbb{E} \Bigg( exp \Bigg[ it \sqrt{m_n} \Bigg( \frac{1}{\binom{n}{k_n}} \sum_{i=1}^{(n,k_n)} \bigg( \randkern(S_i) - \mathbb{E}_{\omega} \randkern(S_i) \bigg) \\
&\hspace{50mm} + \frac{1}{\binom{n}{k_n}} \sum_{i=1}^{(n,k_n)} \bigg( \mathbb{E}_{\omega} \randkern(S_i) - \theta_{k_n}^{*} \bigg) \Bigg) \Bigg] \phi_{n,k_n,m_n}^{(M)}(t) \Bigg). \\
\end{align*}

\noindent Now, note that since we are in the case where $\alpha > 0$, $m_n = O(n) \ll (n,k_n)$ and thus, by the previous result in the case where $\alpha = 0$, the first term converges to 0 and we have

\begin{align*}
\lim_{n \rightarrow \infty} \phi_{n,k_n,m_n}(t) &= \mathbb{E} \Bigg( \lim_{n \rightarrow \infty} exp \bigg[ it \sqrt{m_n} \; \frac{1}{\binom{n}{k_n}} \sum_{i=1}^{(n,k_n)} \bigg( \mathbb{E}_{\omega} \randkern(S_i) - \theta_{k_n}^{*} \bigg) \bigg] \hspace{2mm} \lim_{n \rightarrow \infty} \phi_{n,k_n,m_n}^{(M)}(t) \Bigg) \\
&= \mathbb{E} \Bigg( \lim_{n \rightarrow \infty} exp \bigg[ it \sqrt{m_n} \; U_{\omega; n,k_n,m_n}^{*} \bigg] \hspace{2mm} \lim_{n \rightarrow \infty} \phi_{n,k_n,m_n}^{(M)}(t) \Bigg) \\
\end{align*}

\noindent so that by exactly the same arguments as in the proof of Theorem \ref{subbaggingthm}

\[
\lim_{n \rightarrow \infty} \phi_{n,k_n,m_n}(t) = exp \Bigg[ -t^2 \Big( \frac{k_{n}^{2}}{\alpha} \zeta_{1,k_n}^{*}  + \zeta_{k_n,k_n}^{*} \Big) / 2 \Bigg] \\
\]

\noindent which is the characteristic function of a Normal distribution with mean 0 and variance $\frac{k_{n}^{2}}{\alpha} \zeta_{1,k_{n}}^{*}  + \zeta_{k_n,k_n}^{*}$.  Further, when $\alpha = \infty$, the variance reduces to $\zeta_{k_n,k_n}^{*}$, as desired.   $\blacksquare$

\newpage
\appendix
\section*{Appendix B.}
\label{AppendixB}


\subsection*{Crossed Design Random Forests}

Typically, each tree in a random forest is built according to an independently selected randomization parameter $\randparam$.  Alternatively, for each $\randparam$, we could choose to build an entire set of trees, one for each of the $m_n$ subsamples, so that if $\Omega_n$ randomization parameters are used, a total of $\Omega_n \times m_n$ trees are built.  We could then write the prediction at $\predpoint$ generated by this alternative random forest estimator as

\begin{equation}
\label{AltRF}
\frac{1}{\Omega_n} \frac{1}{m_n} \sum_{(i)} \sum_{(j)} \treefn_{\predpoint,k_n} ((\bm{X}_{i_1},Y_{i_1}), ..., (\bm{X}_{i_{k_n}},Y_{i_{k_n}}); \; \randparam_j)
\end{equation}

\noindent where here, we explicitly treat $\randparam$ as an input to the function.  Statistics of the form in (\ref{AltRF}) are referred to as \emph{two-sample} or \emph{generalized} U-statistics and similar results regarding asymptotic normality have been established for fixed-rank kernels; see \cite{Lee1990} or \cite{vandervaart} for details.

We mention this as a possible alternative only because the resulting estimator takes the well established form of a generalized U-statistic.  Readers familiar with U-statistics may be more comfortable with this approach than with the random kernel approach that more closely resembles the type of random forests used in practice.  However, since this formulation strays from Breiman's original procedure and is far more computationally intensive, we consider only the random kernel version described in Section \ref{subsec:rf}.

\newpage
\appendix
\section*{Appendix C.}
\label{AppendixC}



The algorithm for estimating $\Sigma_{1,k_n}$  as needed for the hypothesis testing procedure is given below.

\begin{algorithm}
\begin{algorithmic}

\FOR {$i$ in 1 to $n_{\tilde{\bm{z}}}$}

\STATE Select initial fixed point $\tilde{\bm{z}}^{(i)}$

\FOR {$j$ in 1 to $n_{MC}$}

	\STATE Select subsample $\mathcal{S}_{\tilde{\bm{z}}^{(i)},j}$ of size $k_n$ from training set that includes $\tilde{\bm{z}}^{(i)}$
	\STATE Build \emph{full} tree using subsample $\mathcal{S}_{\tilde{\bm{z}}^{(i)},j}$
	\STATE Build \emph{reduced} tree using subsample $\mathcal{S}_{\tilde{\bm{z}}^{(i)},j}$ utilizing only reduced feature space
	\STATE Use both full tree and reduced tree to predict at each test point
	\STATE Record difference in predictions
	
\ENDFOR

\STATE Record average of the $n_{MC}$ differences in predictions 

\ENDFOR

\STATE Compute the covariance of the $n_{\tilde{\bm{z}}}$ averages
\end{algorithmic}
\caption{$\Sigma_{1,k_n}$ Estimation Procedure}
\label{algo:Sigma1Est}
\end{algorithm}

\newpage
\appendix
\section*{Appendix D.}
\label{AppendixD}


\subsection*{Distribution of Subbagged Predictions using an Internal Variance Estimate}

Here we examine the distribution of predictions when the ensemble is built according to the internal variance estimation method described in Algorithm \ref{algo:internal}.  Since we are only interested in the distribution of predictions, we omit the steps in Algorithm \ref{algo:internal} for estimating the variance parameters and use the same estimates as in the external case above.  For both the SLR case with $n = 1000, k = 60$ and the MARS case with $n = 1000, k = 75$, we use $n_{\tilde{\bm{z}}} = 50$ and $n_{MC} = 250$ to produce a total of $m = 12500$ predictions in each ensemble.  A total of 250 ensembles were built and the resulting histograms with estimated normal densities overlaid are shown in Figure \ref{fig:internal} below.  We see that these distributions appear to be the same as when the subsamples are selected uniformly at random, as in the external variance estimation method.

\vspace{10mm}

\begin{figure}[H]
  \centering
  \includegraphics[scale=0.4]{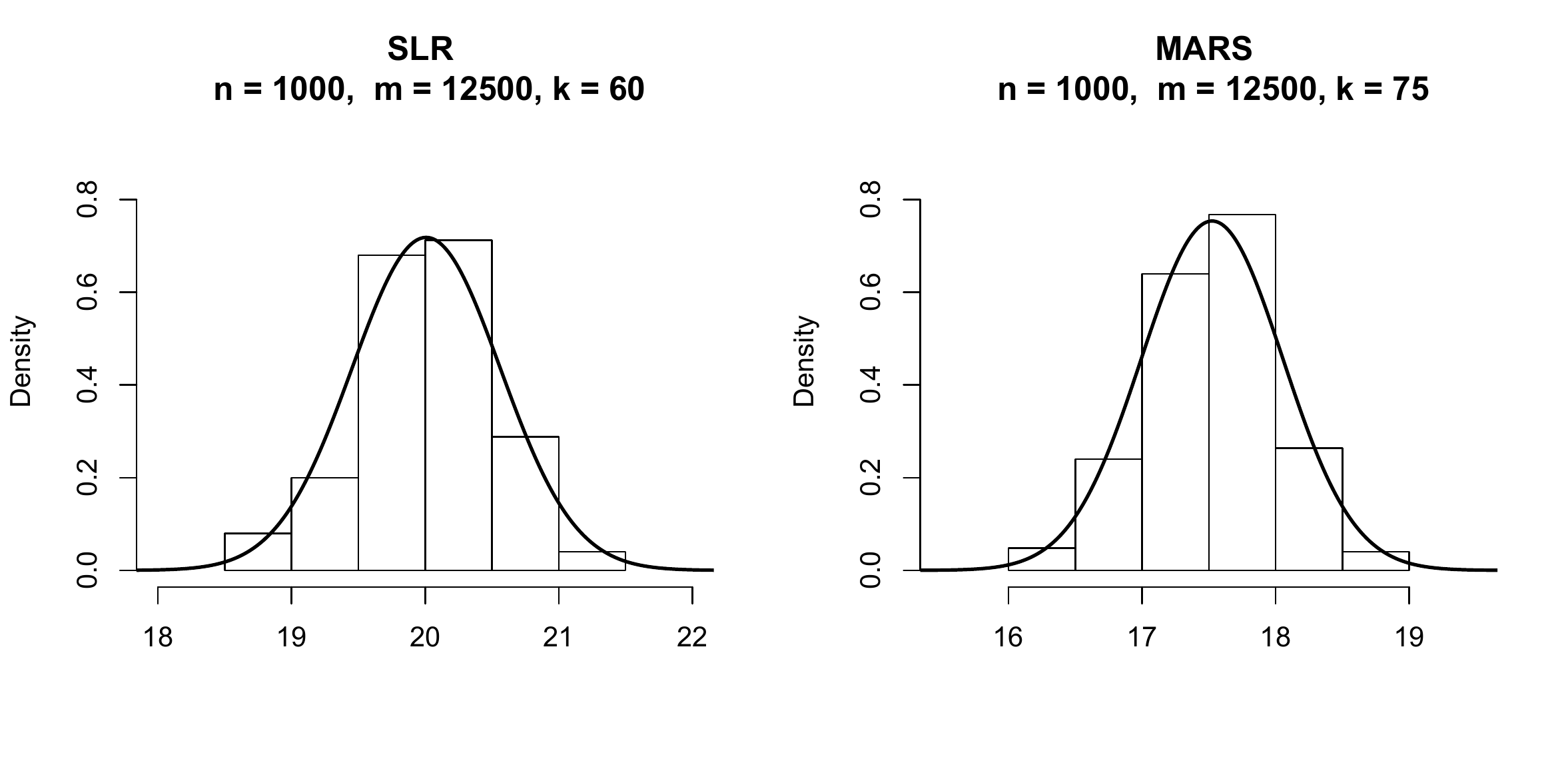}
  \caption{\label{fig:internal} Histograms of subbagged predictions with using an internal estimate of variance.  Predictions are made at $x_1=10$ in the SLR case and at $x_1= \cdots = x_5 = 0.5$ in the MARS case.}
\end{figure}

\end{document}